\documentclass[conference]{IEEEtran}

\usepackage{amsmath,amssymb}   % 数学
\usepackage{graphicx}          % \includegraphics
\usepackage{booktabs}          % \toprule \midrule \bottomrule
\usepackage{multirow}
\usepackage[ruled,vlined,linesnumbered]{algorithm2e} 
\usepackage{bm}
\usepackage[table]{xcolor}
\usepackage{cite}              % IEEE 推荐的数字引用压缩，[1]–[3]
\usepackage[hidelinks]{hyperref}  % 关键：去掉引用/链接的框
\usepackage{orcidlink}

\begin{document}
\bstctlcite{FADTI:BSTcontrol}
\title{FADTI: Fourier and Attention Driven Diffusion for Multivariate Time Series Imputation}

\author{
\IEEEauthorblockN{
Runze Li\textsuperscript{1},
Hanchen Wang\textsuperscript{2},
Wenjie Zhang\textsuperscript{1},
Binghao Li\textsuperscript{1},
Yu Zhang\textsuperscript{1},
Xuemin Lin\textsuperscript{3},
and Ying Zhang\textsuperscript{2}\textsuperscript{*}}
\IEEEauthorblockA{
\textsuperscript{1}University of New South Wales, Australia
\textsuperscript{2}University of Technology Sydney, Australia\\
\textsuperscript{3}The Chinese University of Hong Kong, Shenzhen, China\\
\{runze.li1, wenjie.zhang, binghao.li, m.yuzhang\}@unsw.edu.au
\{hanchen.wang, ying.zhang\}@uts.edu.au; \\
xueminlin@cuhk.edu.hk}
}

\maketitle
\begingroup
\renewcommand{\thefootnote}{}
\footnotetext{
\textsuperscript{*}Corresponding Author.
}
\endgroup

\begin{abstract}
Multivariate time series imputation is fundamental in applications such as healthcare, traffic forecasting, and biological modeling, where sensor failures and irregular sampling lead to pervasive missing values.
Existing Transformer- and diffusion-based imputers achieve strong performance, but they often rely mainly on time-domain modeling and lack adaptive spectral bias for recovering structured temporal gaps.
We propose FADTI, a Fourier- and attention-driven diffusion framework for multivariate time series imputation.
FADTI introduces a Fourier Bias Projection (FBP) module that injects learnable frequency-aware bias into intermediate hidden states during denoising.
It projects intermediate hidden states onto Fourier bases, avoiding direct spectral estimation from masked or zero-filled inputs.
With DFT, STFT, and FSST instantiations, FBP captures global periodicity, localized time--frequency variations, and non-stationary oscillatory patterns.
By coupling FBP with self-attention and gated convolution, FADTI integrates frequency-domain guidance, temporal dependency modeling, and probabilistic denoising in a unified framework.
Experiments on multiple benchmarks, including a new biological imputation benchmark, 
show that FADTI improves accuracy, uncertainty estimation, and sampling efficiency, especially under high missing rates and structured missing patterns.
% show that FADTI improves imputation accuracy and uncertainty estimation while maintaining competitive sampling efficiency, especially under high missing rates and structured missing patterns.
Code is available at https://github.com/RazeenLI/FADTI
% https://anonymous.4open.science/r/FADTI-15C9
\end{abstract}


\section{Introduction}

Multivariate time series imputation (MTSI) is essential in healthcare, finance, Internet of Things (IoT), and transportation, where incomplete observations commonly arise from sensor failures, communication losses, irregular sampling, system downtime, and privacy constraints~\cite{cit:cudre2020mind, cit:economic2008jushan, cit:iot2023xiao, cit:transportation2023yongshun, cit:medical2012ibrahim, cit:medical2025linglong}. 
Such missing values can severely degrade downstream tasks such as forecasting, risk assessment, and decision making, especially when missingness forms long gaps or irregular patterns.

Recent MTSI studies have improved temporal dependency modeling with attention-based architectures and uncertainty estimation with generative models.
% Transformer-based models, such as SAITS and its variants~\cite{cit:saits2023wenjie, cit:dsttn2024xusheng, cit:lai2024rectsi, cit:attention2025cizheng, cit:li2023missing}, capture long-range dependencies by modeling global context across time steps and variables. 
Transformer-based models, such as SAITS and its variants~\cite{cit:saits2023wenjie, cit:dsttn2024xusheng, cit:lai2024rectsi, cit:attention2025cizheng}, capture long-range dependencies by modeling global context across time steps and variables. 
Diffusion-based models, including CSDI, MTSCI, and OVDM~\cite{cit:csdi2021yusuke, cit:mtsci2024jianping, cit:wang2023observed}, further improve imputation by learning a conditional denoising process over partially observed sequences. 
However, severe missingness creates a coupled challenge: the model must infer uncertain missing values while preserving long-term temporal structure. 
Purely time-domain models often lack explicit constraints on periodicity, trend continuity, and spectral consistency, making it difficult to recover long-range or repetitive patterns under sparse observations.

Frequency-domain representations are well suited to describing periodicity, trend continuity, and long-range variation.
Recent studies, including TimesNet, FGTI, PSW-I, and TSLANet~\cite{cit:timesnet2023haixu, cit:fgti2024xin, wang2025optimal, cit:tslanet2024emadeldeen}, have introduced frequency-aware modeling into time series forecasting and classification. 
Nevertheless, directly transferring these techniques to imputation is nontrivial because the input sequence is only partially observed. 
A naive Fourier projection of masked or zero-filled sequences implicitly treats missing positions as real signal values, which distorts the estimated spectrum and makes frequency features sensitive to the filling strategy.
\begin{figure}[!t]
    \centering
    \includegraphics[width=0.95\columnwidth]{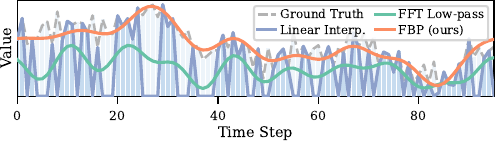}
    \vspace{-4mm}
    \caption{
    Effect of missingness on reconstruction from corrupted inputs.
    }
    \vspace{-5mm}
    \label{fig:freq_visual}
\end{figure}

% \noindent
\noindent\textbf{Example.}
Figure~\ref{fig:freq_visual} highlights how missing values distort frequency-based reconstruction from corrupted inputs.
As shown by the blue curve, simple filling can introduce sharp artificial fluctuations when observations are sparse or irregular.
FFT-based low-pass reconstruction is also unreliable: the green curve departs from the true trend indicated by the gray dashed curve, because its spectrum is estimated from an incomplete sequence.
It therefore follows missingness-induced local variations rather than the underlying smooth trend.
Thus, frequency features computed directly from corrupted inputs can be dominated by missingness and filling artifacts rather than the underlying temporal dynamics.

Motivated by these observations, we propose FADTI, a Fourier and attention driven diffusion framework for MTSI.
FADTI follows a conditional diffusion pipeline to model the distribution of missing values under ambiguous contexts.
In the denoising network, self-attention captures long-range temporal and cross-variable dependencies, while gated convolution captures local temporal patterns with lower computational cost.
By coupling these components in a single denoising process, FADTI produces uncertainty-aware and temporally coherent imputations under sparse or irregular missingness.

To address this frequency distortion, FADTI introduces Fourier Bias Projection (FBP), which injects learnable frequency-aware bias into the denoising network without directly estimating spectra from masked inputs.
During reverse diffusion, FBP projects intermediate denoising features onto Fourier bases and uses the resulting bias to encourage the predicted residual to preserve smooth trends and periodic structures.
The FBP curve in Figure~\ref{fig:freq_visual} illustrates the intended behavior of our adaptive frequency bias, which better preserves the trend and amplitude variations than directly applying Fourier projection to corrupted inputs.
The module can be instantiated with different Fourier projections: the Discrete Fourier Transform (DFT) provides a global frequency representation over the whole sequence~\cite{cit:fbp2024runze, cit:dft1965james}, Short-Time Fourier Transform (STFT) applies windowed Fourier analysis to capture time-varying local spectra~\cite{cit:stft2022sudhakar}, and Fourier Synchrosqueezing Transform (FSST) sharpens non-stationary time-frequency structures through frequency reassignment~\cite{cit:fsst2013Gaurav}.
% This design allows FBP to capture both global and time-varying spectral structures.
This design allows FBP to capture both global and time-varying spectra.

The main contributions of this work are as follows:
\begin{itemize}
\item We propose FADTI, a frequency-aware conditional diffusion framework for MTSI, where Fourier-domain inductive bias is integrated into the denoising process for robust imputation under challenging missing patterns.

\item We design a flexible FBP module with DFT, STFT, and FSST instantiations, enabling adaptive modeling of stationary and non-stationary temporal structures without directly relying on corrupted spectra from masked inputs.

\item We incorporate a lightweight local modeling module into the temporal encoder to complement attention-based global dependencies with lower computational overhead.

\item We conduct extensive experiments on multiple benchmarks, including a new biological imputation benchmark for complex non-stationary environments, demonstrating consistent improvements in accuracy and robustness.
\end{itemize}
\section{Related Work}

\textbf{Multivariate Time Series Imputation.}
Early methods, including statistical imputation, regression, K-nearest neighbors, and matrix factorization, are efficient but rely heavily on local similarity or low-rank assumptions, making them less effective for high-dimensional, nonstationary sequences with irregular missingness~\cite{cit:mice2011stef, cit:knn2019ching, cit:knnjoin2023nimish}.
Recent neural methods learn temporal and cross-variable dependencies directly from data. 
RNN- and CNN-based models exploit sequential states or local temporal filters, but they can struggle with long-range dependencies or require carefully designed receptive fields~\cite{cit:brits2018wei, cit:chen2024laplacian}. 
Graph-based and GNN-based methods model structural or inter-variable relations, although their performance often depends on graph construction and may become costly for large systems~\cite{cit:gnn2022andrea, cit:subgraphsurvey2026runze}.
Transformer-based methods, including SAITS and ImputeFormer, use self-attention to capture global temporal context, but their attention cost still grows rapidly with sequence length and variable dimension~\cite{cit:saits2023wenjie, cit:imputeformer2024tong}.

Since sparse observations make imputation inherently uncertain, generative models have been widely explored for MTSI.
VAEs and GANs can model data distributions and generate plausible missing values, but they may suffer from over-smoothed reconstructions or unstable training~\cite{cit:gpvae2020vincent, cit:vae2022ahmad, cit:gan2018yonghong, cit:naomi2019yukai}.
Diffusion models have become competitive for MTSI by learning conditional denoising over partially observed sequences~\cite{cit:csdi2021yusuke, cit:survey2024yiyuan}. 
CSDI conditions reverse diffusion on observed values, but its Transformer-based denoising network incurs high computational cost.
Subsequent diffusion-based methods improve efficiency, bridge formulations, spatiotemporal conditioning, or spectral guidance~\cite{cit:diffusion2023juan, cit:csbi2023yu, cit:sadi2024zongyu, cit:mtsci2024jianping, cit:pristi2023mingzhe, cit:fgti2024xin}. 
Despite these advances, existing methods often use frequency information as auxiliary guidance or external constraints rather than as a learnable bias throughout reverse diffusion.

\textbf{Frequency-domain Modeling.}
Frequency-domain analysis is useful for characterizing periodicity, trend changes, and long-range dependencies in time series. 
The discrete Fourier transform provides a compact global spectral representation, while the short-time Fourier transform introduces local windows to describe time-varying frequency components~\cite{cit:stft2022sudhakar, cit:fourier1977allen}. 
More flexible transforms, such as FrFT and FSST, can better characterize certain nonstationary signals by adjusting time--frequency concentration~\cite{cit:frft2020jun, cit:frft1980victor, cit:sst2011ingrid, cit:fsst2013Gaurav}. 
These techniques have been explored in forecasting, classification, and time series representation learning, but their use in diffusion-based imputation remains limited.
In this work, we incorporate DFT, STFT, and FSST into a learnable Fourier Bias Projection module, enabling the denoising network to exploit both global periodic patterns and localized nonstationary cues under long-range gaps and severe missingness.

\begin{figure*}[t]
    \centering
    \includegraphics[width=\textwidth]{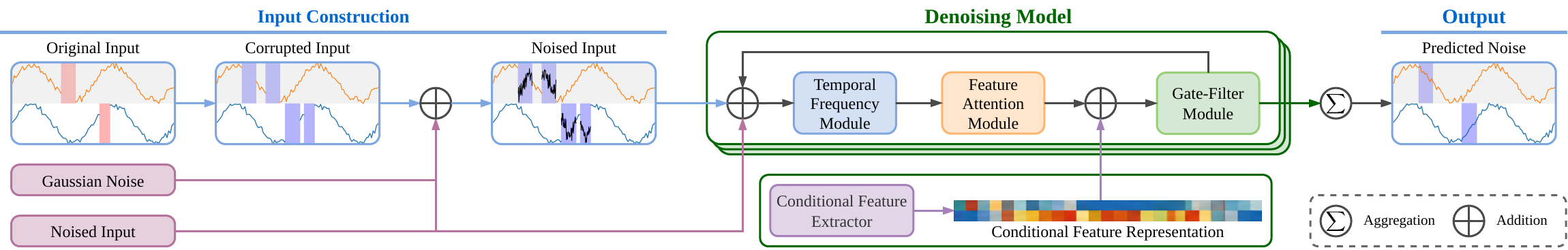}
    \caption{
    Overview of the FADTI framework. 
    The model performs denoising-based imputation by combining conditional feature extraction, temporal--frequency modeling, feature attention, and gated filtering to predict and remove noise from corrupted inputs.
    }
    \label{fig:fadti}
\end{figure*}

\section{Preliminaries}\label{sec:preliminaries}

\subsection{Problem Definition}\label{sec:problem-definition}

\noindent \textit{Multivariate Time Series Imputation (MTSI).}
Let $\bm{x}_0 \in \mathbb{R}^{T \times D}$ denote an underlying complete multivariate time series with $T$ time steps and $D$ variables. 
Let $\bm{m} \in \{0,1\}^{T \times D}$ be the observation mask, where $\bm{m}[t,d]=1$ indicates that $\bm{x}_0[t,d]$ is observed and $\bm{m}[t,d]=0$ otherwise. 
\begin{equation}
    \bm{x}_0^{\mathrm{co}} = \bm{m} \odot \bm{x}_0, 
    \qquad
    \bm{x}_0^{\mathrm{ta}} = (1-\bm{m}) \odot \bm{x}_0,
\end{equation}
where $\odot$ denotes element-wise multiplication, and both tensors have the same shape as $\bm{x}_0$.
The goal of MTSI is to infer $\bm{x}_0^{\mathrm{ta}}$ from $\bm{x}_0^{\mathrm{co}}$ and $\bm{m}$.  
For probabilistic imputation, we model
\begin{equation}
    p_\theta(\bm{x}_0^{\mathrm{ta}} \mid \bm{x}_0^{\mathrm{co}}, \bm{m}),
\end{equation}
from which a point estimate $\hat{\bm{x}}_0^{\mathrm{ta}}$ can be obtained for evaluation. 
Compared with univariate imputation, MTSI must jointly model temporal dependencies, cross-variable correlations, and dimension-specific missingness patterns.

\subsection{Diffusion Models for Imputation}

Diffusion models define a forward process that gradually perturbs a data sample $\bm{x}_0$ into Gaussian noise. 
Given a noise schedule $\{\alpha_s\}_{s=1}^{S}$ and $\bar{\alpha}_s=\prod_{i=1}^{s}\alpha_i$, the noisy sample at diffusion step $s$ can be directly sampled as
\begin{equation}
    \bm{x}_s
    =
    \sqrt{\bar{\alpha}_s}\bm{x}_0
    +
    \sqrt{1-\bar{\alpha}_s}\boldsymbol{\epsilon},
    \quad
    \boldsymbol{\epsilon}\sim\mathcal{N}(\bm{0},\bm{I}).
    \label{eq:noisy_function}
\end{equation}

The reverse process learns to denoise $\bm{x}_s$ step by step:
\begin{equation}
    p_\theta(\bm{x}_{0:S})
    =
    p(\bm{x}_S)
    \prod_{s=1}^{S}
    p_\theta(\bm{x}_{s-1}\mid \bm{x}_s),
    \quad
    p(\bm{x}_S)=\mathcal{N}(\bm{0},\bm{I}).
\end{equation}
Following DDPM~\cite{cit:ddpm2020ho}, the denoiser is trained to predict the injected noise with the standard objective
\begin{equation}
    \mathcal{L}_{\mathrm{diff}}
    =
    \mathbb{E}_{\bm{x}_0,\boldsymbol{\epsilon},s}
    \left[
    \left\|
    \boldsymbol{\epsilon}
    -
    \boldsymbol{\epsilon}_\theta(\bm{x}_s,s)
    \right\|_2^2
    \right].
    \label{eq:diffusion_loss}
\end{equation}

For imputation, the reverse process is conditioned on the observed values and the mask:
\begin{equation}
    p_\theta(\bm{x}_0^{\mathrm{ta}}
    \mid
    \bm{x}_0^{\mathrm{co}},\bm{m})
    =
    p(\bm{x}_S^{\mathrm{ta}})
    \prod_{s=1}^{S}
    p_\theta(
    \bm{x}_{s-1}^{\mathrm{ta}}
    \mid
    \bm{x}_s^{\mathrm{ta}},
    \bm{x}_0^{\mathrm{co}},
    \bm{m}
    ).
    \label{eq:cond_repro}
\end{equation}
The corresponding denoiser is written as
\begin{equation}
    \boldsymbol{\epsilon}_\theta
    =
    \boldsymbol{\epsilon}_\theta(
    \bm{x}_s^{\mathrm{ta}},
    \bm{x}_0^{\mathrm{co}},
    \bm{m},
    s
    ),
\end{equation}
which provides the basis for the conditional denoising architecture introduced in the next section.

\subsection{Fourier-domain Notation}
\label{sec:fourier-notation}

For a multivariate time series $\bm{x}\in\mathbb{R}^{T\times D}$, we use $\mathcal{F}_{\nu}(\cdot)$ to denote a temporal spectral transform applied along the time dimension, where $\nu \in \{\mathrm{DFT}, \mathrm{STFT}, \mathrm{FSST}\}$. 
DFT captures global frequency components, STFT describes localized time-frequency patterns, and Fourier Synchrosqueezing Transform (FSST) further sharpens time-frequency representations through energy reassignment. 
We denote the complex-valued transformed representation as
\begin{equation}
    \tilde{\bm{x}}^{\nu} = \mathcal{F}_{\nu}(\bm{x}).
\end{equation}
For DFT, $\tilde{\bm{x}}^{\nu}$ is indexed by frequency and variable, while STFT and FSST additionally introduce a temporal-frame index.
These spectral operators are used to construct the Fourier Bias Projection module introduced in the next section, where their outputs are converted into learnable denoising biases.

\section{Methodology}
\label{sec:methodology}

\subsection{Overview}
\label{sec:method-overview}

Given an incomplete multivariate time series 
$\bm{X}\in\mathbb{R}^{B\times T\times D}$ and its observation mask 
$\bm{M}\in\{0,1\}^{B\times T\times D}$, where $B$, $T$, and $D$ denote batch size, time steps, and variables, our goal is to infer the missing values conditioned on the observed entries. 
We propose FADTI, a conditional diffusion framework for multivariate time series imputation.
As shown in Figure~\ref{fig:fadti}, FADTI first constructs corrupted and noisy inputs for diffusion, and extracts conditional features from the observed values and mask.
The denoising backbone is built from stacked residual blocks, where each block combines temporal--frequency modeling, feature attention, and gated filtering.
In the temporal--frequency path, FBP converts intermediate hidden states into learnable frequency-aware representations, which are further modeled by temporal attention or gated temporal convolution.
The resulting features are then passed through feature attention and gated filtering to produce the noise prediction used for diffusion denoising.

\begin{figure*}[t]
    \centering
    \includegraphics[width=\textwidth]{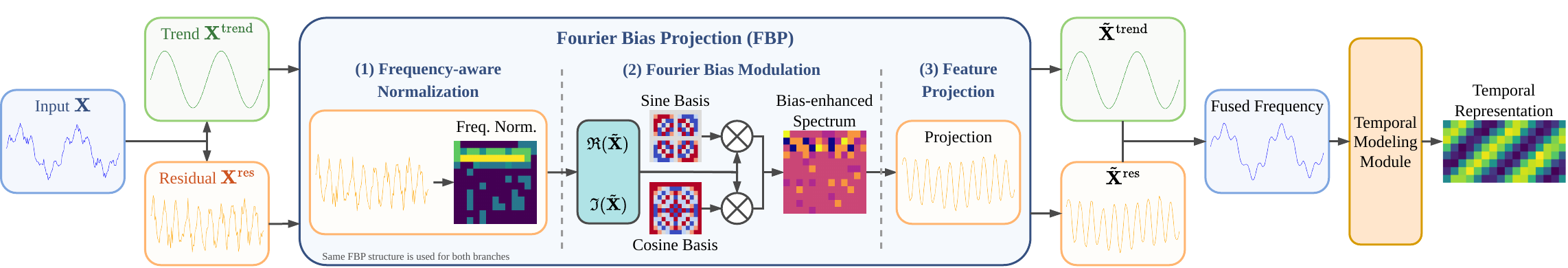}
    \caption{
    Temporal--frequency path with Fourier Bias Projection. 
    FBP generates frequency-aware representations from intermediate hidden features and feeds them to temporal attention or gated convolution within each residual denoising block.
    }
    \label{fig:temporal_frequency_module}
\end{figure*}

\subsection{Fourier Bias Projection}
\label{sec:fbp}

To introduce frequency-domain inductive bias into the residual denoising flow, we design the \textbf{Fourier Bias Projection (FBP)} module. 
Unlike direct Fourier projection on masked or zero-filled inputs, which can distort spectra by treating missing positions as observations, FBP injects learnable frequency-aware bias via Fourier-basis projection of intermediate hidden features. 
This guides temporal attention or gated convolution toward smooth, periodic, and localized non-stationary patterns.

Given an intermediate representation 
$\bm{H}\in\mathbb{R}^{B\times D\times C\times T}$, the spectral operators defined in Section~\ref{sec:fourier-notation} are applied channel-wise along the temporal dimension. 
FBP first separates $\bm{H}$ into a smooth trend and a high-frequency residual:
\begin{equation}
\begin{split}
\bm{H}^{\mathrm{trend}} 
&= \mathrm{AvgPool1D}(\mathrm{Pad}(\bm{H}), K_d), \\
\bm{H}^{\mathrm{res}} 
&= \bm{H} - \bm{H}^{\mathrm{trend}},
\end{split}
\label{eq:temporal_decomposition}
\end{equation}
where $K_d=24$ denotes the smoothing kernel size fixed across all datasets.
The trend branch captures slowly varying temporal structure, while the residual branch preserves local fluctuations and short-term temporal deviations.

For compact notation, let $\bm{Z}$ denote either $\bm{H}^{\mathrm{trend}}$ or $\bm{H}^{\mathrm{res}}$. 
For each component, we apply a spectral transform $\mathcal{F}_{\nu}(\cdot)$:
\begin{equation}
    \tilde{\bm{Z}}^{\nu} = \mathcal{F}_{\nu}(\bm{Z}),
    \qquad
    \nu \in \{\mathrm{DFT}, \mathrm{STFT}, \mathrm{FSST}\}.
    \label{eq:spectral_transform_method}
\end{equation}
Let $F$ be the number of retained frequency bins. 
For DFT, we set $L=T$ and broadcast each retained frequency coefficient over temporal positions before basis projection; for STFT and FSST, $L$ denotes the number of time--frequency frames.
We define fixed cosine and sine bases 
$\bm{b}^{\cos}, \bm{b}^{\sin}\in\mathbb{R}^{F\times L}$ as
\begin{equation}
\bm{b}^{\cos}[f,\ell] = 
\cos\left(\frac{2\pi f\ell}{L}\right),
\quad
\bm{b}^{\sin}[f,\ell] = 
-\sin\left(\frac{2\pi f\ell}{L}\right).
\label{eq:fourier_bases}
\end{equation}
The real and imaginary parts of the spectral coefficients are projected onto these bases:
\begin{equation}
\tilde{\bm{Z}}^{\mathrm{proj}}
=
\Re(\tilde{\bm{Z}}^{\nu}) \odot \bm{b}^{\cos}
+
\Im(\tilde{\bm{Z}}^{\nu}) \odot \bm{b}^{\sin},
\label{eq:fourier_projection}
\end{equation}
where $\odot$ denotes element-wise multiplication with broadcasting over batch, variable, and channel dimensions. 
We then map the result back to the temporal feature space:
\begin{equation}
    \bm{Z}^{\mathrm{freq}}
    =
    \mathrm{Linear}
    \left(
    \mathrm{Dropout}
    \left(
    \mathrm{Flatten}
    \left(
    \tilde{\bm{Z}}^{\mathrm{proj}}
    \right)
    \right)
    \right).
    \label{eq:fbp_output}
\end{equation}
Applying Eq.~\eqref{eq:fbp_output} to the trend and residual branches gives $\bm{H}^{\mathrm{trend,freq}}$ and $\bm{H}^{\mathrm{res,freq}}$, which are fused as
\begin{equation}
    \bm{H}^{\mathrm{freq}}
    =
    \bm{H}^{\mathrm{trend,freq}}
    +
    \bm{H}^{\mathrm{res,freq}},
    \label{eq:fbp_fusion}
\end{equation}
The branches and residual blocks use separate projections before temporal attention or gated convolution; only the Fourier bases are fixed.
By projecting the real and imaginary components onto paired cosine and sine bases, FBP retains the magnitude--phase information of spectral coefficients and converts it into a learnable real-valued frequency-aware bias.

\subsection{Spectral Instantiations of FBP}
\label{subsec:spectral_instantiations}

We instantiate $\mathcal{F}_{\nu}(\cdot)$ with DFT, STFT, and FSST, which provide complementary spectral biases for global periodicity, localized variations, and non-stationary oscillations.

\subsubsection{DFT-based FBP}
DFT maps each temporal trajectory of $\bm{Z}$ to global Fourier coefficients:
\begin{equation}
    \tilde{\bm{Z}}^{\mathrm{DFT}}[...,f]
    =
    \sum_{t=0}^{T-1}
    \bm{Z}[...,t] e^{-2\pi i f t/T},
    \label{eq:dft_method}
\end{equation}
where $[\cdots]$ denotes the omitted batch, variable, and channel indices. 
DFT captures global trends and stable periodic components, but cannot localize time-varying frequencies.

\subsubsection{STFT-based FBP}
STFT introduces localized spectral analysis with a window function $w(\cdot)$:
\begin{equation}
    \tilde{\bm{Z}}^{\mathrm{STFT}}[..., \ell,f]
    =
    \sum_{n=0}^{K_f-1}
    \bm{Z}[..., \ell r+n]\, w[n] e^{-2\pi i f n/K_f},
    \label{eq:stft_method}
\end{equation}
where $K_f$ is the window length, $r$ is the stride, and $\ell$ indexes the temporal frame. 
This representation captures localized time--frequency variations and is more suitable for sequences whose frequency content changes over time.

\subsubsection{FSST-based FBP}
The FSST-based variant starts from STFT coefficients and applies synchrosqueezing-style frequency reassignment to obtain a sharper time--frequency representation. 
It reassigns spectral energy according to an instantaneous-frequency estimate:
\begin{equation}
    \tilde{\bm{Z}}^{\mathrm{FSST}}[..., \tau,\xi]
    =
    \int
    \tilde{\bm{Z}}^{\mathrm{STFT}}[..., \tau,f]
    \delta
    \left(
    \xi - \omega(\tau,f)
    \right)
    df,
    \label{eq:fsst_method}
\end{equation}
where $\omega(\tau,f)$ denotes the instantaneous-frequency estimate, $\xi$ is the reassigned frequency bin, and $[\cdots]$ denotes the omitted batch, variable, and channel indices. 
In implementation, DFT is computed by real FFT, and STFT uses Hann windows of length $K_f=\lfloor2T/3\rfloor$ with 50\% overlap. 
For FSST, the integral reassignment is approximated by discretizing $\omega(\tau,f)$ to valid frequency bins and accumulating the corresponding STFT coefficients. 
This synchrosqueezing step reduces time--frequency smearing and enhances non-stationary oscillations.

\subsubsection{Complexity}
Let $N_{\mathrm{se}}=B\times D\times C$ denote the number of univariate temporal trajectories processed in parallel, and let $T_o$ denote the output length of the projection layer. 
For DFT-based FBP, we have $F=\lfloor T/2 \rfloor+1$ frequency bins. 
The real FFT costs $\mathcal{O}(T\log T)$ per trajectory, the Fourier-basis expansion costs $\mathcal{O}(FT)$, and the flatten--linear projection from $FT$ coefficients to $T_o$ outputs costs $\mathcal{O}(FTT_o)$. 
% Thus, the time and memory complexities are
Thus, the corresponding time and memory complexities are
\begin{equation}
    \mathcal{O}
    \left(
    N_{\mathrm{se}}
    \left(
    T\log T + FT + FTT_o
    \right)
    \right),
    \;
    \mathcal{O}
    \left(
    N_{\mathrm{se}}FT
    \right).
    \label{eq:dft_complexity}
\end{equation}

For STFT-based FBP, each trajectory is divided into $L$ frames with window length $K_f$, where $F=\lfloor K_f/2 \rfloor+1$. 
The windowed FFTs cost $\mathcal{O}(LK_f\log K_f)$, the basis expansion costs $\mathcal{O}(FL)$, and the flatten--linear projection from $FL$ coefficients to $T_o$ outputs costs $\mathcal{O}(FLT_o)$. 
Thus, the corresponding time and memory complexities are
\begin{equation}
    \mathcal{O}
    \left(
    N_{\mathrm{se}}
    \left(
    LK_f\log K_f + FL + FLT_o
    \right)
    \right),
    \;
    \mathcal{O}
    \left(
    N_{\mathrm{se}}FL
    \right).
    \label{eq:stft_complexity}
\end{equation}

For FSST-based FBP, the STFT computation is followed by instantaneous-frequency estimation and synchrosqueezing reassignment. 
These additional operations are linear in the number of time--frequency coefficients, i.e., $\mathcal{O}(FL)$ per trajectory. 
Therefore, the time and memory complexities are
\begin{equation}
    \mathcal{O}
    \left(
    N_{\mathrm{se}}
    \left(
    LK_f\log K_f + 2FL + FLT_o
    \right)
    \right),
        \;
    \mathcal{O}
    \left(
    N_{\mathrm{se}}FL
    \right),
    \label{eq:fsst_complexity}
\end{equation}
with a larger constant factor than STFT-based FBP due to instantaneous-frequency estimation and reassignment.

\subsection{Temporal and Feature Modeling}
\label{sec:temporal-feature}

After FBP produces the frequency-aware representation $\bm{H}^{\mathrm{freq}}$, the temporal--frequency path processes it using either temporal self-attention or gated temporal convolution. 
For the attention-based variant, each variable is treated as a sequence along time. 
The input is viewed as $B\times D$ temporal sequences of length $T$ with $C$ channels, and multi-head self-attention is applied along the temporal dimension:
\begin{equation}
    \bm{H}^{\mathrm{temp}}
    =
    \mathrm{Transformer}_{t}(\bm{H}^{\mathrm{freq}}),
    \label{eq:temporal_attention}
\end{equation}
where queries, keys, and values are obtained from linear projections of $\bm{H}^{\mathrm{freq}}$ to model long-range temporal dependencies.

As an efficient alternative, we use a gated dilated temporal convolutional network:
\begin{equation}
    \bm{H}^{\mathrm{temp}}
    =
    \tanh(\mathrm{Conv}_{t,f}(\bm{H}^{\mathrm{freq}}))
    \odot
    \sigma(\mathrm{Conv}_{t,g}(\bm{H}^{\mathrm{freq}})),
    \label{eq:gated_tcn}
\end{equation}
where $\mathrm{Conv}_{t,f}$ and $\mathrm{Conv}_{t,g}$ are dilated temporal convolutions for efficient local pattern modeling.
To model cross-variable dependencies, we apply feature attention over the $D$ variables:
\begin{equation}
    \bm{H}^{\mathrm{feat}}
    =
    \mathrm{Transformer}_{d}(\bm{H}^{\mathrm{temp}}).
    \label{eq:feature_attention}
\end{equation}

Gated filtering then updates the residual representation:
\begin{equation}
    \bm{G}^{r}
    =
    \tanh(\mathrm{Conv}_{r,f}(\bm{H}^{\mathrm{feat}}))
    \odot
    \sigma(\mathrm{Conv}_{r,g}(\bm{H}^{\mathrm{feat}})).
    \label{eq:gated_filter}
\end{equation}
The residual and skip outputs are computed as
\begin{equation}
    \bm{H}^{r+1}
    =
    \bm{H}^{r}
    +
    \mathrm{Conv}_{\mathrm{res}}(\bm{G}^{r}),
    \qquad
    \bm{S}^{r}
    =
    \mathrm{Conv}_{\mathrm{skip}}(\bm{G}^{r}).
    \label{eq:residual_update}
\end{equation}
Here, $\bm{H}^{r}$ and $\bm{H}^{r+1}$ denote the input and output states of the $r$-th residual block, and $\bm{S}^{r}$ is collected by the final projection head for diffusion noise prediction.

\subsection{Training}
\label{sec:training}

We train the model with a masked denoising score matching objective. 
Since ground truth is available only at originally observed entries, supervision is restricted to these entries.

\subsubsection{Mask construction}
During training, we sample a conditional mask $\bm{M}^{\mathrm{co}}$ from the observation mask $\bm{M}$:
\begin{equation}
    \bm{M}^{\mathrm{co}}
    =
    \bm{M}\odot\bm{R},
    \qquad
    \bm{R}_{btd}\sim\mathrm{Bernoulli}(p),
    \label{eq:conditional_mask}
\end{equation}
and define the target mask as
\begin{equation}
    \bm{M}^{\mathrm{ta}}
    =
    \bm{M}-\bm{M}^{\mathrm{co}}.
    \label{eq:target_mask}
\end{equation}
Here, $\bm{M}^{\mathrm{co}}$ contains values revealed to the model as conditions, while $\bm{M}^{\mathrm{ta}}$ contains originally observed values that are intentionally hidden for supervision. 
Entries with $\bm{M}=0$ are excluded from the loss because they lack ground truth.

\subsubsection{Noise corruption and denoising}
At each iteration, we sample $s\sim\mathcal{U}(1,S)$ and corrupt the input by
\begin{equation}
    \bm{X}_{s}
    =
    \sqrt{\bar{\alpha}_s}\bm{X}
    +
    \sqrt{1-\bar{\alpha}_s}\boldsymbol{\epsilon},
    \qquad
    \boldsymbol{\epsilon}\sim\mathcal{N}(\bm{0},\bm{I}).
    \label{eq:noise_corruption}
\end{equation}
We combine clean conditions and noisy targets as
\begin{equation}
    \bm{X}_s^{\mathrm{in}}
    =
    \mathrm{Concat}
    \left(
    \bm{M}^{\mathrm{co}}\odot\bm{X},
    (1-\bm{M}^{\mathrm{co}})\odot\bm{X}_{s}
    \right).
    \label{eq:denoising_input}
\end{equation}
Original missing entries are excluded from supervision by $\bm{M}^{\mathrm{ta}}$, and the denoiser predicts the injected noise:
\begin{equation}
    \hat{\boldsymbol{\epsilon}}
    =
    \boldsymbol{\epsilon}_{\theta}
    \left(
    \bm{X}_s^{\mathrm{in}},
    s,
    \bm{X}^{\mathrm{cf}}
    \right).
    \label{eq:noise_prediction}
\end{equation}
The loss is computed only on the supervised target mask:
\begin{equation}
\mathcal{L}_{\mathrm{train}}
=
\frac{
\sum_{b,t,d}
\bm{M}^{\mathrm{ta}}_{btd}
\left(
\boldsymbol{\epsilon}_{btd}
-
\hat{\boldsymbol{\epsilon}}_{btd}
\right)^2
}{
\sum_{b,t,d}
\bm{M}^{\mathrm{ta}}_{btd}
}.
\label{eq:masked_denoising_loss}
\end{equation}
We further inject side information, including time embeddings, variable embeddings, and mask embeddings, into the denoising network to provide positional and missingness-aware context.

\subsection{Inference}
\label{sec:inference}

At inference time, observed entries are fixed and missing entries are initialized with Gaussian noise. 
Starting from $\bm{X}^{\mathrm{ta}}_{S}\sim\mathcal{N}(\bm{0},\bm{I})$, we iteratively apply the learned reverse process. 
Each DDPM step constructs the input, predicts the noise, and updates the target sample:
\begin{equation}
\begin{gathered}
\begin{aligned}
\bm{X}_{s}^{\mathrm{in}}
&=
\mathrm{Concat}
\left(
\bm{M}\odot\bm{X},
(1-\bm{M})\odot\bm{X}_{s}^{\mathrm{ta}}
\right), \\
\hat{\boldsymbol{\epsilon}}_s
&=
\boldsymbol{\epsilon}_\theta
\left(
\bm{X}_{s}^{\mathrm{in}},
s,
\bm{X}^{\mathrm{cf}}
\right), 
\end{aligned}
\\
\bm{X}_{s-1}^{\mathrm{ta}}
=
\frac{1}{\sqrt{\alpha_s}}
\left(
\bm{X}_{s}^{\mathrm{ta}}
-
\frac{1-\alpha_s}{\sqrt{1-\bar{\alpha}_s}}
\hat{\boldsymbol{\epsilon}}_s
\right)
+
\sigma_s\boldsymbol{\zeta},
\qquad
\boldsymbol{\zeta}\sim\mathcal{N}(\bm{0},\bm{I}).
\end{gathered}
\label{eq:reverse_sampling}
\end{equation}
We repeat the reverse process $K$ times and average the generated target samples:
\begin{equation}
\hat{\bm{X}}
=
\bm{M}\odot\bm{X}
+
(1-\bm{M})
\odot
\frac{1}{K}
\sum_{k=1}^{K}
(\bm{X}_{0}^{\mathrm{ta}})_k .
\label{eq:sample_average}
\end{equation}
\section{Experiments}
\label{sec:experiments}

We evaluate FADTI on multivariate time series imputation across diverse datasets, missing patterns, and missing rates. 
The experiments examine imputation accuracy, uncertainty quality, robustness, sampling efficiency, architectural and spectral variants, noise choices, and scenario studies.

\begin{table}[tbp]
\centering
\caption{Summary of datasets used in the experiments.}
\label{tab:datasets}
\resizebox{\columnwidth}{!}{
\begin{tabular}{llcccc}
\toprule
Dataset & Domain & \# of Features & \# of Steps & Interval & Missing (\%) \\
\midrule
ETT     & Energy      & 7      & 96  & 15 min & 0    \\
Weather & Meteorology & 21     & 144 & 10 min & 0.02 \\
METR-LA & Traffic     & 207    & 288 & 5 min  & 8.6  \\
E. coli & Biology     & 7      & 185 & 5 min  & 0    \\
\bottomrule
\end{tabular}
}
\end{table}

\subsection{Experimental Setup}
\label{sec:experimental_setup}

\begin{table*}[!t]
\small
\centering
\caption{MAE ($\downarrow$) of different imputation methods under point- and time-wise missing patterns on the ETT, Weather, and E. coli datasets.}
\label{tab:mae}
\resizebox{\textwidth}{!}{
\begin{tabular}{@{}l *{12}{c}@{}}
\toprule
\multirow{3}{*}{Method} & 
\multicolumn{4}{c}{ETT} & \multicolumn{4}{c}{Weather} & \multicolumn{4}{c}{E. coli} \\
\cmidrule(lr){2-5} \cmidrule(lr){6-9} \cmidrule(lr){10-13} 
& \multicolumn{2}{c}{Point} & \multicolumn{2}{c}{Time} & \multicolumn{2}{c}{Point} & \multicolumn{2}{c}{Time} & \multicolumn{2}{c}{Point} & \multicolumn{2}{c}{Time} \\
\cmidrule(lr){2-3} \cmidrule(lr){4-5} \cmidrule(lr){6-7} \cmidrule(lr){8-9} \cmidrule(lr){10-11} \cmidrule(lr){12-13}
& 0.1 & 0.5 & 0.1 & 0.5 & 0.1 & 0.5 & 0.1 & 0.5 & 0.1 & 0.5 & 0.1 & 0.5 \\
\midrule

Mean & 2.485$_{\scriptscriptstyle \pm 0.405}$ & 2.487$_{\scriptscriptstyle \pm 0.413}$ & 2.477$_{\scriptscriptstyle \pm 0.410}$ & 2.483$_{\scriptscriptstyle \pm 0.404}$ & 81.890$_{\scriptscriptstyle \pm 22.348}$ & 81.145$_{\scriptscriptstyle \pm 21.774}$ & 81.166$_{\scriptscriptstyle \pm 21.310}$ & 81.173$_{\scriptscriptstyle \pm 21.560}$ & 190.875$_{\scriptscriptstyle \pm 3.337}$ & 191.444$_{\scriptscriptstyle \pm 3.754}$ & 190.532$_{\scriptscriptstyle \pm 3.679}$ & 191.120$_{\scriptscriptstyle \pm 3.245}$ \\
Median & 2.399$_{\scriptscriptstyle \pm 0.362}$ & 2.401$_{\scriptscriptstyle \pm 0.373}$ & 2.383$_{\scriptscriptstyle \pm 0.360}$ & 2.395$_{\scriptscriptstyle \pm 0.352}$ & 68.883$_{\scriptscriptstyle \pm 21.668}$ & 68.269$_{\scriptscriptstyle \pm 21.175}$ & 68.224$_{\scriptscriptstyle \pm 20.596}$ & 68.030$_{\scriptscriptstyle \pm 21.289}$ & 168.183$_{\scriptscriptstyle \pm 4.067}$ & 169.279$_{\scriptscriptstyle \pm 4.722}$ & 168.607$_{\scriptscriptstyle \pm 5.560}$ & 168.661$_{\scriptscriptstyle \pm 4.456}$ \\
KNN & 1.037$_{\scriptscriptstyle \pm 0.045}$ & 1.445$_{\scriptscriptstyle \pm 0.142}$ & 1.450$_{\scriptscriptstyle \pm 0.142}$ & 1.501$_{\scriptscriptstyle \pm 0.147}$ & 23.428$_{\scriptscriptstyle \pm 8.212}$ & 28.084$_{\scriptscriptstyle \pm 9.636}$ & 24.663$_{\scriptscriptstyle \pm 8.106}$ & 27.358$_{\scriptscriptstyle \pm 9.295}$ & 137.771$_{\scriptscriptstyle \pm 1.889}$ & 144.068$_{\scriptscriptstyle \pm 1.369}$ & 137.051$_{\scriptscriptstyle \pm 1.198}$ & 141.542$_{\scriptscriptstyle \pm 0.853}$ \\
BRITS & 0.317$_{\scriptscriptstyle \pm 0.070}$ & 0.558$_{\scriptscriptstyle \pm 0.091}$ & \underline{0.505$_{\scriptscriptstyle \pm 0.112}$} & 0.758$_{\scriptscriptstyle \pm 0.166}$ & 9.119$_{\scriptscriptstyle \pm 3.542}$ & 12.416$_{\scriptscriptstyle \pm 3.881}$ & \underline{12.262$_{\scriptscriptstyle \pm 3.308}$} & \underline{16.976$_{\scriptscriptstyle \pm 5.080}$} & 20.488$_{\scriptscriptstyle \pm 2.023}$ & 31.619$_{\scriptscriptstyle \pm 15.651}$ & 24.176$_{\scriptscriptstyle \pm 4.229}$ & 46.691$_{\scriptscriptstyle \pm 3.878}$ \\
SAITS & 0.939$_{\scriptscriptstyle \pm 0.255}$ & 1.239$_{\scriptscriptstyle \pm 0.647}$ & 2.199$_{\scriptscriptstyle \pm 0.645}$ & 2.182$_{\scriptscriptstyle \pm 0.718}$ & 30.231$_{\scriptscriptstyle \pm 3.365}$ & 70.782$_{\scriptscriptstyle \pm 5.464}$ & 156.038$_{\scriptscriptstyle \pm 14.629}$ & 149.107$_{\scriptscriptstyle \pm 17.515}$ & 63.263$_{\scriptscriptstyle \pm 12.793}$ & 78.286$_{\scriptscriptstyle \pm 30.717}$ & 299.771$_{\scriptscriptstyle \pm 93.125}$ & 361.723$_{\scriptscriptstyle \pm 76.027}$ \\
TimesNet & 0.599$_{\scriptscriptstyle \pm 0.240}$ & 0.553$_{\scriptscriptstyle \pm 0.206}$ & 0.593$_{\scriptscriptstyle \pm 0.228}$ & \underline{0.562$_{\scriptscriptstyle \pm 0.226}$} & 13.435$_{\scriptscriptstyle \pm 5.251}$ & 12.935$_{\scriptscriptstyle \pm 4.653}$ & \underline{13.578$_{\scriptscriptstyle \pm 4.078}$} & \underline{12.997$_{\scriptscriptstyle \pm 4.393}$} & 17.038$_{\scriptscriptstyle \pm 3.050}$ & 18.939$_{\scriptscriptstyle \pm 7.682}$ & 21.429$_{\scriptscriptstyle \pm 10.014}$ & 28.751$_{\scriptscriptstyle \pm 10.070}$ \\
TimeMixer & 2.358$_{\scriptscriptstyle \pm 0.392}$ & 2.331$_{\scriptscriptstyle \pm 0.363}$ & 2.344$_{\scriptscriptstyle \pm 0.379}$ & 2.346$_{\scriptscriptstyle \pm 0.378}$ & 114.180$_{\scriptscriptstyle \pm 22.617}$ & 116.147$_{\scriptscriptstyle \pm 12.892}$ & 117.844$_{\scriptscriptstyle \pm 19.074}$ & 116.006$_{\scriptscriptstyle \pm 11.099}$ & -- & -- & -- & -- \\
TimeMixer++ & 1.026$_{\scriptscriptstyle \pm 0.357}$ & 0.951$_{\scriptscriptstyle \pm 0.200}$ & 1.234$_{\scriptscriptstyle \pm 0.262}$ & 1.168$_{\scriptscriptstyle \pm 0.675}$ & 50.108$_{\scriptscriptstyle \pm 46.546}$ & 74.575$_{\scriptscriptstyle \pm 44.588}$ & 30.214$_{\scriptscriptstyle \pm 14.160}$ & 29.760$_{\scriptscriptstyle \pm 10.756}$ & -- & -- & -- & -- \\
CSDI & \underline{0.240$_{\scriptscriptstyle \pm 0.027}$} & \underline{0.327$_{\scriptscriptstyle \pm 0.039}$} & \underline{0.452$_{\scriptscriptstyle \pm 0.118}$} & \underline{0.617$_{\scriptscriptstyle \pm 0.201}$} & \underline{6.347$_{\scriptscriptstyle \pm 2.283}$} & \underline{9.184$_{\scriptscriptstyle \pm 3.195}$} & 41.433$_{\scriptscriptstyle \pm 6.030}$ & 41.443$_{\scriptscriptstyle \pm 6.673}$ & \underline{7.013$_{\scriptscriptstyle \pm 1.486}$} & \underline{10.876$_{\scriptscriptstyle \pm 1.876}$} & \underline{20.138$_{\scriptscriptstyle \pm 3.514}$} & \underline{21.546$_{\scriptscriptstyle \pm 3.377}$} \\
MTSCI & \underline{0.248$_{\scriptscriptstyle \pm 0.033}$} & \underline{0.332$_{\scriptscriptstyle \pm 0.040}$} & 0.844$_{\scriptscriptstyle \pm 0.084}$ & 0.722$_{\scriptscriptstyle \pm 0.090}$ & \textbf{\underline{5.574$_{\scriptscriptstyle \pm 2.234}$}} & \underline{8.199$_{\scriptscriptstyle \pm 2.914}$} & 41.568$_{\scriptscriptstyle \pm 5.832}$ & 41.165$_{\scriptscriptstyle \pm 6.061}$ & \underline{7.392$_{\scriptscriptstyle \pm 1.406}$} & \underline{10.238$_{\scriptscriptstyle \pm 2.004}$} & \underline{17.580$_{\scriptscriptstyle \pm 3.024}$} & \textbf{\underline{19.449$_{\scriptscriptstyle \pm 3.515}$}} \\
SSDTS & 0.961$_{\scriptscriptstyle \pm 0.373}$ & 1.150$_{\scriptscriptstyle \pm 0.410}$ & 1.419$_{\scriptscriptstyle \pm 0.609}$ & 1.259$_{\scriptscriptstyle \pm 0.465}$ & 258.168$_{\scriptscriptstyle \pm 10.446}$ & 172.130$_{\scriptscriptstyle \pm 31.478}$ & 257.734$_{\scriptscriptstyle \pm 9.705}$ & 224.053$_{\scriptscriptstyle \pm 27.922}$ & 67.932$_{\scriptscriptstyle \pm 13.804}$ & 77.754$_{\scriptscriptstyle \pm 3.078}$ & 78.203$_{\scriptscriptstyle \pm 10.377}$ & 80.154$_{\scriptscriptstyle \pm 0.397}$ \\
\midrule FADTI & \textbf{\underline{0.197$_{\scriptscriptstyle \pm 0.011}$}} & \textbf{\underline{0.283$_{\scriptscriptstyle \pm 0.031}$}} & \textbf{\underline{0.329$_{\scriptscriptstyle \pm 0.048}$}} & \textbf{\underline{0.377$_{\scriptscriptstyle \pm 0.057}$}} & \underline{6.946$_{\scriptscriptstyle \pm 4.563}$} & \textbf{\underline{6.327$_{\scriptscriptstyle \pm 3.097}$}} & \textbf{\underline{9.279$_{\scriptscriptstyle \pm 6.743}$}} & \textbf{\underline{8.271$_{\scriptscriptstyle \pm 3.978}$}} & \textbf{\underline{5.950$_{\scriptscriptstyle \pm 1.538}$}} & \textbf{\underline{9.438$_{\scriptscriptstyle \pm 1.895}$}} & \textbf{\underline{16.987$_{\scriptscriptstyle \pm 3.070}$}} & \underline{19.493$_{\scriptscriptstyle \pm 3.310}$} \\

\bottomrule
\end{tabular}
}
% \end{tabular*}
% }
\end{table*}

\subsubsection{Datasets}

We evaluate FADTI on three real-world multivariate time series datasets from diverse domains: ETT~\cite{cit:informer2021haoyi}, Weather~\cite{cit:saits2023wenjie}, and E. coli~\cite{cit:ecoli2024jean}.
ETT contains transformer temperature and load readings, while Weather consists of meteorological variables such as temperature and humidity.
METR-LA~\cite{cit:metrla2018yaguang} is used for additional analysis.

The E. coli dataset originates from single-cell bacterial experiments, where cell morphology and gene expression are tracked over time.
We use Set2, which contains 3,456 cell trajectories with 185 time steps each.
From the original 12 variables, we select 7 continuous features covering cell shape and size, fluorescence intensity, image sharpness, and cell count.
We discard chamber-level aggregates and binary intervention indicators, since they either obscure single-cell dynamics or are not continuous imputation targets.
We include E. coli as a challenging biological benchmark to evaluate imputation under noisy and weakly periodic dynamics.

Table~\ref{tab:datasets} summarizes the dataset characteristics.
All datasets are normalized using training-split statistics, and missing masks are generated following the experimental protocol.

\begin{table*}[t]
\small
\centering
\caption{CRPS ($\downarrow$) of different imputation methods under point- and time-wise missing patterns on the ETT, Weather, and E. coli datasets.}
\label{tab:crps}
% \resizebox{\textwidth}{!}{
% \begin{tabular*}{\textwidth}{@{\extracolsep{\fill}} l *{12}{c} @{}}
\resizebox{\textwidth}{!}{
\begin{tabular}{@{}l *{12}{c}@{}}
% \begin{tabular}{lcccccccccccccc}
\toprule
\multirow{3}{*}{Method} & 
\multicolumn{4}{c}{ETT} & \multicolumn{4}{c}{Weather} & \multicolumn{4}{c}{E. coli} \\
\cmidrule(lr){2-5} \cmidrule(lr){6-9} \cmidrule(lr){10-13} 
& \multicolumn{2}{c}{Point} & \multicolumn{2}{c}{Time} & \multicolumn{2}{c}{Point} & \multicolumn{2}{c}{Time} & \multicolumn{2}{c}{Point} & \multicolumn{2}{c}{Time} \\
\cmidrule(lr){2-3} \cmidrule(lr){4-5} \cmidrule(lr){6-7} \cmidrule(lr){8-9} \cmidrule(lr){10-11} \cmidrule(lr){12-13}
& 0.1 & 0.5 & 0.1 & 0.5 & 0.1 & 0.5 & 0.1 & 0.5 & 0.1 & 0.5 & 0.1 & 0.5 \\
\midrule

Mean & 0.532$_{\scriptscriptstyle \pm 0.138}$ & 0.535$_{\scriptscriptstyle \pm 0.135}$ & 0.534$_{\scriptscriptstyle \pm 0.137}$ & 0.535$_{\scriptscriptstyle \pm 0.135}$ & 0.319$_{\scriptscriptstyle \pm 0.075}$ & 0.317$_{\scriptscriptstyle \pm 0.073}$ & 0.317$_{\scriptscriptstyle \pm 0.072}$ & 0.317$_{\scriptscriptstyle \pm 0.073}$ & 0.493$_{\scriptscriptstyle \pm 0.006}$ & 0.493$_{\scriptscriptstyle \pm 0.005}$ & 0.491$_{\scriptscriptstyle \pm 0.004}$ & 0.493$_{\scriptscriptstyle \pm 0.004}$ \\
Median & 0.513$_{\scriptscriptstyle \pm 0.128}$ & 0.515$_{\scriptscriptstyle \pm 0.125}$ & 0.513$_{\scriptscriptstyle \pm 0.125}$ & 0.515$_{\scriptscriptstyle \pm 0.123}$ & 0.281$_{\scriptscriptstyle \pm 0.074}$ & 0.280$_{\scriptscriptstyle \pm 0.072}$ & 0.280$_{\scriptscriptstyle \pm 0.072}$ & 0.279$_{\scriptscriptstyle \pm 0.074}$ & 0.458$_{\scriptscriptstyle \pm 0.006}$ & 0.460$_{\scriptscriptstyle \pm 0.005}$ & 0.458$_{\scriptscriptstyle \pm 0.005}$ & 0.458$_{\scriptscriptstyle \pm 0.004}$ \\
KNN & 0.221$_{\scriptscriptstyle \pm 0.029}$ & 0.311$_{\scriptscriptstyle \pm 0.058}$ & 0.313$_{\scriptscriptstyle \pm 0.059}$ & 0.324$_{\scriptscriptstyle \pm 0.061}$ & 0.091$_{\scriptscriptstyle \pm 0.028}$ & 0.110$_{\scriptscriptstyle \pm 0.033}$ & 0.096$_{\scriptscriptstyle \pm 0.030}$ & 0.107$_{\scriptscriptstyle \pm 0.033}$ & 0.362$_{\scriptscriptstyle \pm 0.005}$ & 0.378$_{\scriptscriptstyle \pm 0.007}$ & 0.361$_{\scriptscriptstyle \pm 0.008}$ & 0.372$_{\scriptscriptstyle \pm 0.008}$ \\
BRITS & 0.066$_{\scriptscriptstyle \pm 0.006}$ & 0.118$_{\scriptscriptstyle \pm 0.009}$ & \underline{0.107$_{\scriptscriptstyle \pm 0.016}$} & 0.160$_{\scriptscriptstyle \pm 0.021}$ & 0.036$_{\scriptscriptstyle \pm 0.013}$ & 0.049$_{\scriptscriptstyle \pm 0.014}$ & \underline{0.049$_{\scriptscriptstyle \pm 0.012}$} & \underline{0.068$_{\scriptscriptstyle \pm 0.019}$} & 0.054$_{\scriptscriptstyle \pm 0.006}$ & 0.084$_{\scriptscriptstyle \pm 0.044}$ & 0.064$_{\scriptscriptstyle \pm 0.012}$ & 0.122$_{\scriptscriptstyle \pm 0.009}$ \\
SAITS & 0.200$_{\scriptscriptstyle \pm 0.050}$ & 0.258$_{\scriptscriptstyle \pm 0.121}$ & 0.471$_{\scriptscriptstyle \pm 0.143}$ & 0.475$_{\scriptscriptstyle \pm 0.182}$ & 0.113$_{\scriptscriptstyle \pm 0.017}$ & 0.263$_{\scriptscriptstyle \pm 0.028}$ & 0.571$_{\scriptscriptstyle \pm 0.039}$ & 0.546$_{\scriptscriptstyle \pm 0.063}$ & 0.155$_{\scriptscriptstyle \pm 0.028}$ & 0.192$_{\scriptscriptstyle \pm 0.074}$ & 0.748$_{\scriptscriptstyle \pm 0.201}$ & 0.910$_{\scriptscriptstyle \pm 0.180}$ \\
TimesNet & 0.122$_{\scriptscriptstyle \pm 0.030}$ & 0.114$_{\scriptscriptstyle \pm 0.026}$ & 0.122$_{\scriptscriptstyle \pm 0.029}$ & 0.116$_{\scriptscriptstyle \pm 0.030}$ & 0.052$_{\scriptscriptstyle \pm 0.018}$ & 0.050$_{\scriptscriptstyle \pm 0.016}$ & \underline{0.053$_{\scriptscriptstyle \pm 0.015}$} & \underline{0.051$_{\scriptscriptstyle \pm 0.016}$} & \underline{0.044$_{\scriptscriptstyle \pm 0.009}$} & 0.052$_{\scriptscriptstyle \pm 0.025}$ & 0.055$_{\scriptscriptstyle \pm 0.024}$ & \underline{0.077$_{\scriptscriptstyle \pm 0.027}$} \\
TimeMixer & 0.494$_{\scriptscriptstyle \pm 0.003}$ & 0.492$_{\scriptscriptstyle \pm 0.004}$ & 0.495$_{\scriptscriptstyle \pm 0.004}$ & 0.496$_{\scriptscriptstyle \pm 0.006}$ & 0.444$_{\scriptscriptstyle \pm 0.079}$ & 0.454$_{\scriptscriptstyle \pm 0.041}$ & 0.459$_{\scriptscriptstyle \pm 0.055}$ & 0.452$_{\scriptscriptstyle \pm 0.036}$ & -- & -- & -- & -- \\
TimeMixer++ & 0.213$_{\scriptscriptstyle \pm 0.053}$ & 0.200$_{\scriptscriptstyle \pm 0.025}$ & 0.264$_{\scriptscriptstyle \pm 0.062}$ & 0.241$_{\scriptscriptstyle \pm 0.107}$ & 0.193$_{\scriptscriptstyle \pm 0.175}$ & 0.301$_{\scriptscriptstyle \pm 0.192}$ & 0.119$_{\scriptscriptstyle \pm 0.057}$ & 0.117$_{\scriptscriptstyle \pm 0.039}$ & -- & -- & -- & -- \\
CSDI & \underline{0.039$_{\scriptscriptstyle \pm 0.004}$} & \underline{0.053$_{\scriptscriptstyle \pm 0.009}$} & \underline{0.077$_{\scriptscriptstyle \pm 0.025}$} & \underline{0.100$_{\scriptscriptstyle \pm 0.037}$} & \underline{0.020$_{\scriptscriptstyle \pm 0.007}$} & \underline{0.029$_{\scriptscriptstyle \pm 0.009}$} & 0.133$_{\scriptscriptstyle \pm 0.016}$ & 0.132$_{\scriptscriptstyle \pm 0.018}$ & \underline{0.015$_{\scriptscriptstyle \pm 0.004}$} & \underline{0.022$_{\scriptscriptstyle \pm 0.005}$} & \underline{0.041$_{\scriptscriptstyle \pm 0.008}$} & \underline{0.044$_{\scriptscriptstyle \pm 0.009}$} \\
MTSCI & \underline{0.040$_{\scriptscriptstyle \pm 0.004}$} & \underline{0.054$_{\scriptscriptstyle \pm 0.008}$} & 0.125$_{\scriptscriptstyle \pm 0.021}$ & \underline{0.113$_{\scriptscriptstyle \pm 0.019}$} & \textbf{\underline{0.018$_{\scriptscriptstyle \pm 0.006}$}} & \underline{0.026$_{\scriptscriptstyle \pm 0.009}$} & 0.155$_{\scriptscriptstyle \pm 0.019}$ & 0.153$_{\scriptscriptstyle \pm 0.019}$ & \underline{0.015$_{\scriptscriptstyle \pm 0.003}$} & \underline{0.021$_{\scriptscriptstyle \pm 0.005}$} & \underline{0.035$_{\scriptscriptstyle \pm 0.007}$} & \textbf{\underline{0.039$_{\scriptscriptstyle \pm 0.009}$}} \\
SSDTS & 0.157$_{\scriptscriptstyle \pm 0.041}$ & 0.202$_{\scriptscriptstyle \pm 0.048}$ & 0.232$_{\scriptscriptstyle \pm 0.069}$ & 0.222$_{\scriptscriptstyle \pm 0.053}$ & 0.981$_{\scriptscriptstyle \pm 0.022}$ & 0.518$_{\scriptscriptstyle \pm 0.091}$ & 0.946$_{\scriptscriptstyle \pm 0.095}$ & 0.670$_{\scriptscriptstyle \pm 0.085}$ & 0.151$_{\scriptscriptstyle \pm 0.026}$ & 0.185$_{\scriptscriptstyle \pm 0.009}$ & 0.174$_{\scriptscriptstyle \pm 0.024}$ & 0.188$_{\scriptscriptstyle \pm 0.007}$ \\
\midrule FADTI & \textbf{\underline{0.032$_{\scriptscriptstyle \pm 0.004}$}} & \textbf{\underline{0.046$_{\scriptscriptstyle \pm 0.005}$}} & \textbf{\underline{0.054$_{\scriptscriptstyle \pm 0.012}$}} & \textbf{\underline{0.062$_{\scriptscriptstyle \pm 0.014}$}} & \underline{0.021$_{\scriptscriptstyle \pm 0.012}$} & \textbf{\underline{0.020$_{\scriptscriptstyle \pm 0.009}$}} & \textbf{\underline{0.029$_{\scriptscriptstyle \pm 0.019}$}} & \textbf{\underline{0.025$_{\scriptscriptstyle \pm 0.012}$}} & \textbf{\underline{0.012$_{\scriptscriptstyle \pm 0.004}$}} & \textbf{\underline{0.019$_{\scriptscriptstyle \pm 0.004}$}} & \textbf{\underline{0.034$_{\scriptscriptstyle \pm 0.007}$}} & \textbf{\underline{0.039$_{\scriptscriptstyle \pm 0.008}$}} \\

\bottomrule
\end{tabular}
}
\end{table*}

\subsubsection{Missing Settings}
We evaluate two missing patterns that simulate common sensor-level failures: \textbf{point-wise missing}, where individual entries are randomly masked across time and variables, and \textbf{time-wise missing}, where selected time steps are masked across all variables.
We consider 10\% and 50\% missing rates, and use identical masks across methods for each dataset, missing pattern, missing rate, and random seed.

\subsubsection{Evaluation Metrics}
MAE and RMSE measure reconstruction accuracy, MAPE evaluates relative error across scales, and CRPS assesses the calibration and sharpness of probabilistic imputations. 
For probabilistic models, CRPS is computed from generated samples;
for deterministic baselines, we treat their point estimates as degenerate distributions.

\subsubsection{Baselines}
We compare FADTI with representative time series imputation baselines:
\begin{itemize}
\raggedright
    \item \textbf{Statistical}: Mean, Median imputation.
    \item \textbf{Distance}: KNN.
    \item \textbf{Neural}: BRITS~\cite{cit:brits2018wei}, SAITS~\cite{cit:saits2023wenjie}, TimesNet~\cite{cit:timesnet2023haixu}, TimeMixer~\cite{cit:timemixer2024shiyu}, and TimeMixer++~\cite{cit:timemixer++2025shiyu}.
    \item \textbf{Diffusion}: CSDI~\cite{cit:csdi2021yusuke}, MTSCI~\cite{cit:mtsci2024jianping} and SSDTS~\cite{cit:ssdts2025Hongfan}.
\end{itemize}
TimesNet and TimeMixer++ are included as non-diffusion models that exploit spectral or multi-scale temporal structure.

\subsubsection{Implementation Protocol}
All experiments are conducted in PyTorch on a single NVIDIA RTX A5000 GPU. 
% For each dataset, missing setting, and random seed, all methods use the same train/validation/test split with a 70\%/10\%/20\% ratio and generated masks.
All methods use the same 70\%/10\%/20\% train/validation/test split and generated masks for each setting.
The five runs correspond to different folds, and we report the mean and standard deviation.
For baselines with official implementations, we use the released code and adapt only the data loader, masking protocol, and input/output interface to our benchmark. 
All baselines are retrained on our datasets rather than copied from previously reported results. 
% Hyperparameters are selected on the validation set following the recommended settings of each method when available and using a common validation search otherwise, with detailed configurations provided in the released code.
Hyperparameters are selected on the validation set using recommended settings when available, with detailed configurations provided in the released code.
Unless otherwise specified, the main FADTI results use DFT-based FBP with temporal self-attention.
For a fair comparison, diffusion-based methods use $S=50$ diffusion steps and generate $K=100$ imputation samples for evaluation.

\begin{table*}[t]
\small
\centering
\caption{RMSE and MAPE ($\downarrow$) of different imputation methods under point- and time-wise missing patterns on the ETT, Weather, and E. coli datasets with 50\% missing ratio.}
\label{tab:rmse_mape}
\resizebox{\textwidth}{!}{
\begin{tabular}{@{}l *{12}{c}@{}}
\toprule
\multirow{3}{*}{Method} & 
\multicolumn{4}{c}{ETT} & \multicolumn{4}{c}{Weather} & \multicolumn{4}{c}{E. coli} \\
\cmidrule(lr){2-5} \cmidrule(lr){6-9} \cmidrule(lr){10-13} 
& \multicolumn{2}{c}{Point} & \multicolumn{2}{c}{Time} & \multicolumn{2}{c}{Point} & \multicolumn{2}{c}{Time} & \multicolumn{2}{c}{Point} & \multicolumn{2}{c}{Time} \\
\cmidrule(lr){2-3} \cmidrule(lr){4-5} \cmidrule(lr){6-7} \cmidrule(lr){8-9} \cmidrule(lr){10-11} \cmidrule(lr){12-13}
& RMSE & MAPE & RMSE & MAPE & RMSE & MAPE & RMSE & MAPE & RMSE & MAPE & RMSE & MAPE \\
\midrule
BRITS & 0.965$_{\scriptscriptstyle \pm 0.292}$ & 4.943$_{\scriptscriptstyle \pm 1.649}$ & 1.300$_{\scriptscriptstyle \pm 0.376}$ & 13.053$_{\scriptscriptstyle \pm 16.295}$ & 98.511$_{\scriptscriptstyle \pm 49.809}$ & \underline{58.985$_{\scriptscriptstyle \pm 14.594}$} & \underline{98.085$_{\scriptscriptstyle \pm 43.688}$} & \underline{98.830$_{\scriptscriptstyle \pm 45.170}$} & 106.228$_{\scriptscriptstyle \pm 32.444}$ & 257.319$_{\scriptscriptstyle \pm 174.984}$ & 129.701$_{\scriptscriptstyle \pm 11.131}$ & 315.851$_{\scriptscriptstyle \pm 65.841}$ \\
SAITS & 1.762$_{\scriptscriptstyle \pm 0.893}$ & 13.130$_{\scriptscriptstyle \pm 9.966}$ & 2.830$_{\scriptscriptstyle \pm 0.829}$ & 13.059$_{\scriptscriptstyle \pm 6.421}$ & 125.580$_{\scriptscriptstyle \pm 16.128}$ & 145.772$_{\scriptscriptstyle \pm 21.119}$ & 281.272$_{\scriptscriptstyle \pm 27.107}$ & 322.303$_{\scriptscriptstyle \pm 81.518}$ & 131.307$_{\scriptscriptstyle \pm 77.081}$ & 545.734$_{\scriptscriptstyle \pm 256.054}$ & 765.427$_{\scriptscriptstyle \pm 120.766}$ & 1431.712$_{\scriptscriptstyle \pm 920.870}$ \\
TimesNet & 0.966$_{\scriptscriptstyle \pm 0.527}$ & 5.753$_{\scriptscriptstyle \pm 3.653}$ & \underline{0.986$_{\scriptscriptstyle \pm 0.578}$} & \underline{8.746$_{\scriptscriptstyle \pm 9.831}$} & \underline{92.576$_{\scriptscriptstyle \pm 46.473}$} & 75.549$_{\scriptscriptstyle \pm 17.050}$ & \underline{92.487$_{\scriptscriptstyle \pm 44.152}$} & \underline{83.892$_{\scriptscriptstyle \pm 21.588}$} & 58.444$_{\scriptscriptstyle \pm 23.867}$ & 140.514$_{\scriptscriptstyle \pm 81.434}$ & \textbf{\underline{84.853$_{\scriptscriptstyle \pm 22.597}$}} & 167.337$_{\scriptscriptstyle \pm 104.454}$ \\
TimeMixer & 3.236$_{\scriptscriptstyle \pm 0.843}$ & \underline{3.140$_{\scriptscriptstyle \pm 0.335}$} & 3.251$_{\scriptscriptstyle \pm 0.874}$ & \textbf{\underline{3.266$_{\scriptscriptstyle \pm 0.481}$}} & 179.092$_{\scriptscriptstyle \pm 16.102}$ & 165.654$_{\scriptscriptstyle \pm 14.435}$ & 173.087$_{\scriptscriptstyle \pm 13.723}$ & 172.899$_{\scriptscriptstyle \pm 17.868}$ & -- & -- & -- & -- \\
TimeMixer++ & 1.734$_{\scriptscriptstyle \pm 0.693}$ & 8.369$_{\scriptscriptstyle \pm 4.014}$ & 2.030$_{\scriptscriptstyle \pm 1.262}$ & 18.127$_{\scriptscriptstyle \pm 25.471}$ & 162.833$_{\scriptscriptstyle \pm 74.099}$ & 240.239$_{\scriptscriptstyle \pm 163.602}$ & 109.648$_{\scriptscriptstyle \pm 47.481}$ & 226.294$_{\scriptscriptstyle \pm 95.798}$ & -- & -- & -- & -- \\
CSDI & \textbf{\underline{0.619$_{\scriptscriptstyle \pm 0.115}$}} & 3.641$_{\scriptscriptstyle \pm 1.891}$ & \underline{1.081$_{\scriptscriptstyle \pm 0.339}$} & 15.368$_{\scriptscriptstyle \pm 22.636}$ & 95.956$_{\scriptscriptstyle \pm 47.385}$ & 67.633$_{\scriptscriptstyle \pm 20.258}$ & 129.377$_{\scriptscriptstyle \pm 39.411}$ & 144.451$_{\scriptscriptstyle \pm 48.822}$ & \underline{56.051$_{\scriptscriptstyle \pm 14.979}$} & \underline{103.915$_{\scriptscriptstyle \pm 75.602}$} & 97.213$_{\scriptscriptstyle \pm 20.859}$ & \underline{162.367$_{\scriptscriptstyle \pm 100.597}$} \\
MTSCI & \underline{0.631$_{\scriptscriptstyle \pm 0.104}$} & \underline{3.501$_{\scriptscriptstyle \pm 1.521}$} & 1.248$_{\scriptscriptstyle \pm 0.120}$ & 19.140$_{\scriptscriptstyle \pm 28.601}$ & \textbf{\underline{91.810$_{\scriptscriptstyle \pm 47.297}$}} & \underline{55.445$_{\scriptscriptstyle \pm 11.297}$} & 127.124$_{\scriptscriptstyle \pm 37.407}$ & 124.210$_{\scriptscriptstyle \pm 16.011}$ & \underline{53.769$_{\scriptscriptstyle \pm 16.044}$} & \textbf{\underline{101.122$_{\scriptscriptstyle \pm 76.439}$}} & \underline{95.317$_{\scriptscriptstyle \pm 20.946}$} & \textbf{\underline{137.851$_{\scriptscriptstyle \pm 80.088}$}} \\
SSDTS & 2.055$_{\scriptscriptstyle \pm 1.114}$ & 5.424$_{\scriptscriptstyle \pm 1.847}$ & 2.160$_{\scriptscriptstyle \pm 1.068}$ & 9.937$_{\scriptscriptstyle \pm 8.774}$ & 278.833$_{\scriptscriptstyle \pm 48.795}$ & 347.885$_{\scriptscriptstyle \pm 71.853}$ & 334.340$_{\scriptscriptstyle \pm 39.885}$ & 442.093$_{\scriptscriptstyle \pm 54.424}$ & 221.761$_{\scriptscriptstyle \pm 11.521}$ & 290.523$_{\scriptscriptstyle \pm 67.246}$ & 220.792$_{\scriptscriptstyle \pm 5.450}$ & 367.596$_{\scriptscriptstyle \pm 110.786}$ \\
\midrule FADTI & \underline{0.636$_{\scriptscriptstyle \pm 0.243}$} & \textbf{\underline{2.657$_{\scriptscriptstyle \pm 0.771}$}} & \textbf{\underline{0.744$_{\scriptscriptstyle \pm 0.192}$}} & \underline{9.737$_{\scriptscriptstyle \pm 14.818}$} & \underline{93.906$_{\scriptscriptstyle \pm 57.177}$} & \textbf{\underline{48.685$_{\scriptscriptstyle \pm 8.656}$}} & \textbf{\underline{89.942$_{\scriptscriptstyle \pm 47.068}$}} & \textbf{\underline{81.335$_{\scriptscriptstyle \pm 45.946}$}} & \textbf{\underline{52.556$_{\scriptscriptstyle \pm 15.632}$}} & \underline{102.484$_{\scriptscriptstyle \pm 79.287}$} & \underline{93.621$_{\scriptscriptstyle \pm 20.950}$} & \underline{146.718$_{\scriptscriptstyle \pm 93.187}$} \\

\bottomrule
\end{tabular}
}
\end{table*}

\subsection{Main Imputation Results}
\label{sec:main_results}

Table~\ref{tab:mae} reports MAE results under both point-wise and time-wise missing patterns. 
Overall, FADTI yields the lowest MAE in most settings and consistently ranks among the top methods. 
In all result tables, bold indicates the best result, underlining marks the top three methods, and ``--'' denotes configurations that could not be executed with the official implementation under our fixed input length.

\subsubsection{Point-wise missing}
Under point-wise missing, FADTI achieves the best or top-three MAE in nearly all settings, with particularly strong gains at the 50\% missing rate. 
The improvements are more pronounced on Weather and E. coli, indicating that combining conditional diffusion with frequency-aware reconstruction is beneficial under severe sparsity.

\subsubsection{Time-wise missing}
Under time-wise missing, FADTI also achieves the lowest MAE in most settings. 
The main exception is E. coli at the 50\% missing rate, where it remains close to the best-performing MTSCI. 
Compared with point-wise missing, several baselines, including SAITS, CSDI, and MTSCI, degrade more under time-wise gaps, especially on Weather and E. coli.
This indicates that frequency-aware bias helps recover temporal gaps that local context alone cannot resolve.
Methods with explicit temporal modeling, such as TimesNet and BRITS, are more competitive under time-wise gaps, highlighting the importance of temporal dependency modeling.
FADTI further improves this setting by coupling temporal modeling with frequency-domain bias.

\subsection{Uncertainty and Robustness}
\label{sec:uncertainty_robustness}

Beyond MAE, we report CRPS, RMSE, and MAPE to evaluate probabilistic quality, large-error sensitivity, and relative error.
For readability, these metrics are reported for learning-based baselines and FADTI; results for simple imputers are omitted since they are consistently weaker in MAE.

\subsubsection{Uncertainty Modeling}
CRPS evaluates the quality of predictive distributions, with lower scores indicating better calibration and sharpness.
Table~\ref{tab:crps} reports CRPS scores under different missing patterns and datasets. 
FADTI achieves the lowest CRPS in most cases and remains among the top three elsewhere, indicating a favorable calibration--sharpness trade-off across datasets and missing patterns.

\subsubsection{Sensitivity to Large Errors}
RMSE penalizes large errors more strongly than MAE and is used to assess large local deviations.
As shown in Table~\ref{tab:rmse_mape}, FADTI is consistently among the strongest methods and often achieves the best RMSE. 
This shows that FADTI reduces large deviations while improving average reconstruction accuracy.

\subsubsection{Relative Error across Scales}
Since MAE and RMSE do not directly reflect relative scale across features, we also report MAPE as a complementary metric. 
As shown in Table~\ref{tab:rmse_mape}, FADTI ranks within the top three in most settings and often achieves the best MAPE, suggesting better preservation of relative scale across variables.

\begin{table*}[t]
\small
\centering
\caption{
MAE ($\downarrow$) of FADTI ablations with different frequency modules and temporal blocks.
}
\label{tab:ablation}
\resizebox{\textwidth}{!}{
\begin{tabular}{@{}l *{12}{c}@{}}
\toprule
\multirow{3}{*}{Method} & 
\multicolumn{4}{c}{ETT} & \multicolumn{4}{c}{Weather} & \multicolumn{4}{c}{METR-LA} \\
\cmidrule(lr){2-5} \cmidrule(lr){6-9} \cmidrule(lr){10-13} 
& \multicolumn{2}{c}{Point} & \multicolumn{2}{c}{Time} & \multicolumn{2}{c}{Point} & \multicolumn{2}{c}{Time} & \multicolumn{2}{c}{Point} & \multicolumn{2}{c}{Time} \\
\cmidrule(lr){2-3} \cmidrule(lr){4-5} \cmidrule(lr){6-7} \cmidrule(lr){8-9} \cmidrule(lr){10-11} \cmidrule(lr){12-13}
& 0.1 & 0.5 & 0.1 & 0.5 & 0.1 & 0.5 & 0.1 & 0.5 & 0.1 & 0.5 & 0.1 & 0.5 \\
\midrule

FFT-LP-Attn & 0.275$_{\scriptscriptstyle \pm 0.071}$ & 0.305$_{\scriptscriptstyle \pm 0.024}$ & 0.413$_{\scriptscriptstyle \pm 0.119}$ & \cellcolor{red!10}{0.644$_{\scriptscriptstyle \pm 0.257}$} & \cellcolor{red!10}{7.446$_{\scriptscriptstyle \pm 3.970}$} & \cellcolor{red!10}{10.919$_{\scriptscriptstyle \pm 4.864}$} & 21.614$_{\scriptscriptstyle \pm 2.903}$ & 35.777$_{\scriptscriptstyle \pm 6.417}$ & 3.518$_{\scriptscriptstyle \pm 0.006}$ & 3.738$_{\scriptscriptstyle \pm 0.079}$ & \cellcolor{red!10}{4.958$_{\scriptscriptstyle \pm 0.656}$} & 4.593$_{\scriptscriptstyle \pm 0.126}$ \\
FFT-LP-Conv & 0.224$_{\scriptscriptstyle \pm 0.006}$ & 0.306$_{\scriptscriptstyle \pm 0.019}$ & 0.344$_{\scriptscriptstyle \pm 0.056}$ & 0.397$_{\scriptscriptstyle \pm 0.058}$ & 7.424$_{\scriptscriptstyle \pm 3.291}$ & 9.084$_{\scriptscriptstyle \pm 3.863}$ & 11.884$_{\scriptscriptstyle \pm 4.492}$ & 12.655$_{\scriptscriptstyle \pm 6.217}$ & 3.564$_{\scriptscriptstyle \pm 0.019}$ & 3.799$_{\scriptscriptstyle \pm 0.057}$ & 4.049$_{\scriptscriptstyle \pm 0.570}$ & \underline{4.207$_{\scriptscriptstyle \pm 0.039}$} \\

\specialrule{0.25pt}{0.2ex}{0.2ex}
BP-only-Attn & \cellcolor{red!10}{0.296$_{\scriptscriptstyle \pm 0.125}$} & 0.296$_{\scriptscriptstyle \pm 0.036}$ & \cellcolor{red!10}{0.635$_{\scriptscriptstyle \pm 0.776}$} & 0.558$_{\scriptscriptstyle \pm 0.247}$ & 5.926$_{\scriptscriptstyle \pm 0.208}$ & 8.591$_{\scriptscriptstyle \pm 0.954}$ & \cellcolor{red!10}{24.802$_{\scriptscriptstyle \pm 2.187}$} & \cellcolor{red!10}{39.812$_{\scriptscriptstyle \pm 4.530}$} & 3.440$_{\scriptscriptstyle \pm 0.076}$ & \underline{3.625$_{\scriptscriptstyle \pm 0.039}$} & 4.101$_{\scriptscriptstyle \pm 0.152}$ & \cellcolor{red!10}{4.637$_{\scriptscriptstyle \pm 0.039}$} \\
BP-only-Conv & 0.287$_{\scriptscriptstyle \pm 0.068}$ & \cellcolor{red!10}{0.479$_{\scriptscriptstyle \pm 0.180}$} & 0.428$_{\scriptscriptstyle \pm 0.167}$ & 0.633$_{\scriptscriptstyle \pm 0.225}$ & 6.253$_{\scriptscriptstyle \pm 0.317}$ & 7.446$_{\scriptscriptstyle \pm 1.382}$ & 9.840$_{\scriptscriptstyle \pm 1.542}$ & 11.462$_{\scriptscriptstyle \pm 1.718}$ & \cellcolor{red!10}{3.599$_{\scriptscriptstyle \pm 0.053}$} & \cellcolor{red!10}{3.801$_{\scriptscriptstyle \pm 0.143}$} & 3.818$_{\scriptscriptstyle \pm 0.279}$ & 4.413$_{\scriptscriptstyle \pm 0.041}$ \\
\specialrule{0.25pt}{0.2ex}{0.2ex}
FBP-FSST-Attn & 0.287$_{\scriptscriptstyle \pm 0.179}$ & 0.401$_{\scriptscriptstyle \pm 0.193}$ & 0.307$_{\scriptscriptstyle \pm 0.092}$ & \underline{0.318$_{\scriptscriptstyle \pm 0.063}$} & 6.111$_{\scriptscriptstyle \pm 0.445}$ & 7.211$_{\scriptscriptstyle \pm 0.941}$ & 9.865$_{\scriptscriptstyle \pm 1.770}$ & 9.647$_{\scriptscriptstyle \pm 1.447}$ & 3.439$_{\scriptscriptstyle \pm 0.049}$ & \textbf{\underline{3.609$_{\scriptscriptstyle \pm 0.089}$}} & 4.280$_{\scriptscriptstyle \pm 0.216}$ & 4.456$_{\scriptscriptstyle \pm 0.044}$ \\
FBP-FSST-Conv & 0.275$_{\scriptscriptstyle \pm 0.143}$ & 0.339$_{\scriptscriptstyle \pm 0.124}$ & 0.361$_{\scriptscriptstyle \pm 0.001}$ & 0.386$_{\scriptscriptstyle \pm 0.063}$ & 5.753$_{\scriptscriptstyle \pm 0.097}$ & 6.996$_{\scriptscriptstyle \pm 0.484}$ & 8.367$_{\scriptscriptstyle \pm 0.256}$ & 9.160$_{\scriptscriptstyle \pm 2.768}$ & 3.510$_{\scriptscriptstyle \pm 0.048}$ & 3.723$_{\scriptscriptstyle \pm 0.031}$ & \underline{3.664$_{\scriptscriptstyle \pm 0.109}$} & 4.228$_{\scriptscriptstyle \pm 0.081}$ \\
FBP-STFT-Attn & \underline{0.204$_{\scriptscriptstyle \pm 0.065}$} & 0.277$_{\scriptscriptstyle \pm 0.030}$ & \underline{0.300$_{\scriptscriptstyle \pm 0.086}$} & 0.321$_{\scriptscriptstyle \pm 0.061}$ & \underline{5.338$_{\scriptscriptstyle \pm 0.876}$} & \underline{6.642$_{\scriptscriptstyle \pm 0.605}$} & \underline{7.818$_{\scriptscriptstyle \pm 1.870}$} & 9.807$_{\scriptscriptstyle \pm 1.261}$ & \underline{3.414$_{\scriptscriptstyle \pm 0.066}$} & \underline{3.649$_{\scriptscriptstyle \pm 0.033}$} & 3.783$_{\scriptscriptstyle \pm 0.183}$ & 4.374$_{\scriptscriptstyle \pm 0.214}$ \\
FBP-STFT-Conv & 0.205$_{\scriptscriptstyle \pm 0.018}$ & \underline{0.269$_{\scriptscriptstyle \pm 0.013}$} & \underline{0.298$_{\scriptscriptstyle \pm 0.088}$} & 0.341$_{\scriptscriptstyle \pm 0.002}$ & \underline{5.004$_{\scriptscriptstyle \pm 0.180}$} & 6.895$_{\scriptscriptstyle \pm 0.279}$ & \textbf{\underline{6.882$_{\scriptscriptstyle \pm 1.453}$}} & \underline{7.671$_{\scriptscriptstyle \pm 1.989}$} & 3.487$_{\scriptscriptstyle \pm 0.125}$ & 3.677$_{\scriptscriptstyle \pm 0.103}$ & \textbf{\underline{3.526$_{\scriptscriptstyle \pm 0.031}$}} & \underline{4.093$_{\scriptscriptstyle \pm 0.034}$} \\
FBP-DFT-Attn & \textbf{\underline{0.193$_{\scriptscriptstyle \pm 0.066}$}} & \textbf{\underline{0.221$_{\scriptscriptstyle \pm 0.041}$}} & 0.332$_{\scriptscriptstyle \pm 0.082}$ & \underline{0.320$_{\scriptscriptstyle \pm 0.003}$} & \textbf{\underline{4.182$_{\scriptscriptstyle \pm 1.149}$}} & \textbf{\underline{5.804$_{\scriptscriptstyle \pm 0.094}$}} & \underline{7.422$_{\scriptscriptstyle \pm 0.073}$} & \textbf{\underline{7.345$_{\scriptscriptstyle \pm 1.702}$}} & \underline{3.414$_{\scriptscriptstyle \pm 0.113}$} & 3.714$_{\scriptscriptstyle \pm 0.075}$ & 3.898$_{\scriptscriptstyle \pm 0.059}$ & 4.292$_{\scriptscriptstyle \pm 0.265}$ \\
FBP-DFT-Conv & \underline{0.198$_{\scriptscriptstyle \pm 0.068}$} & \underline{0.255$_{\scriptscriptstyle \pm 0.085}$} & \textbf{\underline{0.246$_{\scriptscriptstyle \pm 0.039}$}} & \textbf{\underline{0.316$_{\scriptscriptstyle \pm 0.000}$}} & 5.560$_{\scriptscriptstyle \pm 0.480}$ & \underline{5.866$_{\scriptscriptstyle \pm 0.323}$} & 11.234$_{\scriptscriptstyle \pm 6.108}$ & \underline{8.191$_{\scriptscriptstyle \pm 0.160}$} & \textbf{\underline{3.394$_{\scriptscriptstyle \pm 0.031}$}} & 3.738$_{\scriptscriptstyle \pm 0.067}$ & \underline{3.701$_{\scriptscriptstyle \pm 0.235}$} & \textbf{\underline{4.039$_{\scriptscriptstyle \pm 0.069}$}} \\

\bottomrule
\end{tabular}
}
\end{table*}

\subsection{Ablation Study: Frequency and Temporal Modules}
\label{sec:ablation}

To evaluate the contributions of FBP and the temporal backbone, we compare ten variants in Table~\ref{tab:ablation}: FFT-LP removes FBP, BP-only keeps back-projection without frequency projection, and FBP-DFT/STFT/FSST instantiate the full FBP module with different frequency representations. 
All variants use the same diffusion framework and training protocol.

As shown in Table~\ref{tab:ablation}, FFT-LP variants are generally unstable and often much worse, while BP-only improves over them in several cases, showing that back-projection is useful.
However, BP-only is still consistently outperformed by the full FBP variants in most settings, especially on ETT and Weather.
This confirms that BP alone is insufficient, and that coupling BP with frequency-domain projection is the key factor behind the performance gains.
Among the full FBP variants, DFT-based models perform strongly on ETT and Weather, suggesting that global spectral information is effective for relatively regular temporal patterns.
STFT- and FSST-based variants become more competitive under time-wise missing settings, where localized or sharper time-frequency representations can better recover time-wise gaps.
The choice between attention and gated convolution has a smaller, more dataset-dependent effect than the frequency representation.

\subsection{Runtime, Memory, and Sampling Efficiency} \label{sec:runtime_memory}

To assess computational efficiency, we report inference time and peak GPU memory in Table~\ref{tab:efficiency}.
All methods are evaluated under the same hardware and batch size.
Inference time is measured on the test set after model loading and warm-up, and peak memory is recorded using CUDA memory statistics during test-time imputation. 
For diffusion-based models, inference cost is measured under the single-sample setting with $K=1$.
FADTI is more expensive than deterministic imputers due to iterative denoising and spectral projection, but it achieves better imputation accuracy and probabilistic quality with fewer generated samples.

\begin{table}[t]
\centering
\caption{Efficiency comparison on ETT under point-wise 10\% missing.}
\label{tab:efficiency}
\begin{tabular}{lccc}
\toprule
Method & Inference Time$_{K=1}$ (s) & Memory (GB) & Parameters (M) \\
\midrule
BRITS & 2.219$_{\scriptscriptstyle \pm 0.209}$ & 0.025$_{\scriptscriptstyle \pm 0.000}$ & 0.11 \\
SAITS & 0.571$_{\scriptscriptstyle \pm 0.066}$ & 0.053$_{\scriptscriptstyle \pm 0.000}$ & 2.12 \\
TimesNet & 0.595$_{\scriptscriptstyle \pm 0.059}$ & 0.029$_{\scriptscriptstyle \pm 0.000}$ & 0.59 \\
TimeMixer & 0.746$_{\scriptscriptstyle \pm 0.332}$ & 0.033$_{\scriptscriptstyle \pm 0.000}$ & 0.09 \\
TimeMixer++ & 2.051$_{\scriptscriptstyle \pm 0.317}$ & 0.066$_{\scriptscriptstyle \pm 0.000}$ & 1.77 \\
CSDI & 5.700$_{\scriptscriptstyle \pm 1.017}$ & 0.162$_{\scriptscriptstyle \pm 0.000}$ & 0.41 \\
MTSCI & 5.969$_{\scriptscriptstyle \pm 0.321}$ & 0.088$_{\scriptscriptstyle \pm 0.000}$ & 0.39 \\
SSDTS & 92.635$_{\scriptscriptstyle \pm 4.427}$ & 0.120$_{\scriptscriptstyle \pm 0.001}$ & 12.04 \\
FADTI & 12.256$_{\scriptscriptstyle \pm 0.627}$ & 0.576$_{\scriptscriptstyle \pm 0.000}$ & 4.06 \\
\bottomrule
\end{tabular}
\end{table}

\begin{figure}[!t]
    \centering
    \includegraphics[width=\columnwidth]{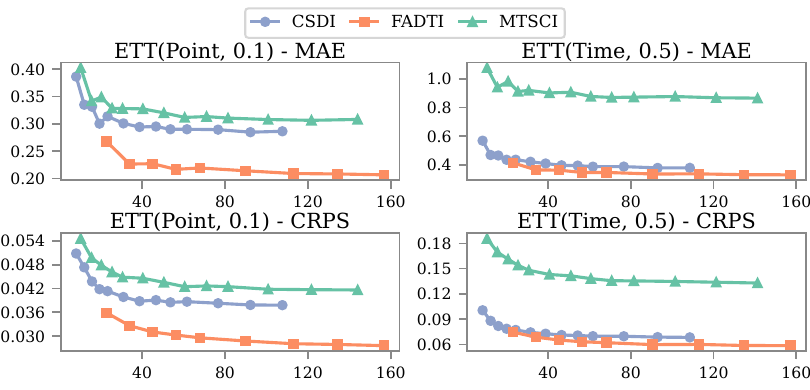}
    \caption{
    MAE and CRPS versus inference time on ETT under point-wise 10\% and time-wise 50\% missing as the sampling budget $K$ varies.
    }
    \label{fig:time_vs_metrics_ett}
\end{figure}

\begin{figure}[!t]
    \centering
    \includegraphics[width=\columnwidth]{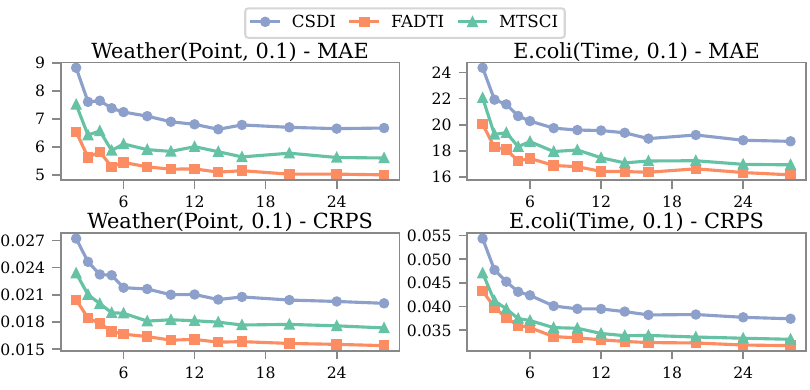}
    \caption{
    MAE and CRPS versus sampling budget $K$ on Weather under point-wise 10\% missing and E. coli under time-wise 10\% missing.
    }
    \label{fig:nsample_vs_metrics_all}
\end{figure}

We further evaluate the sampling efficiency of FADTI against CSDI and MTSCI by varying the number of generated samples $K$. 
Given the substantially higher $K=1$ inference cost of SSDTS in Table~\ref{tab:efficiency}, the sampling-efficiency curves focus on CSDI, MTSCI, and FADTI for readability, under the same evaluation protocol.
% Increasing $K$ generally reduces error at the cost of longer inference time, but the marginal gain becomes small after $K=20$. 
Increasing $K$ generally reduces error but yields diminishing gains after $K=20$.
Across all tested sampling budgets on ETT, FADTI achieves the best MAE and CRPS among the compared methods.
Notably, FADTI with $K=2$ already attains lower MAE and CRPS than CSDI and MTSCI with $K=28$, indicating that FADTI is not only more accurate but also more sample-efficient.
Figure~\ref{fig:nsample_vs_metrics_all} further compares FADTI with CSDI and MTSCI on Weather under point-wise 10\% missing and E. coli under time-wise 10\% missing.
Across both datasets and all tested sampling budgets, FADTI attains the lowest MAE and CRPS, yielding consistent gains over CSDI and MTSCI.
These results suggest that FADTI reaches favorable accuracy with fewer samples, reducing the sampling burden across datasets.

\begin{figure}[!t]
    \centering
    \includegraphics[width=\columnwidth]{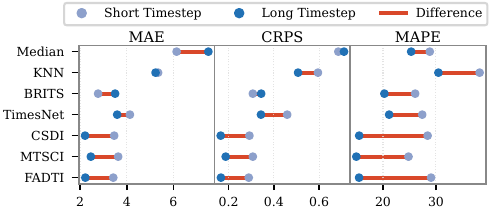}
    \caption{
    Performance comparison on METR-LA under point-wise 10\% missing. Red segments indicate the short--long horizon gap $\Delta$ for each metric.
    }
    \label{fig:metrla_case}
\end{figure}

\subsection{Case Study on METR-LA}
\label{sec:metrla_case_study}

To better understand model behavior under different horizon and dimensionality settings, we conduct a case study on METR-LA with two configurations: (1) all 207 sensors with a 24-step horizon, and (2) 9 sensors with the full 288-step horizon. 
Figure~\ref{fig:metrla_case} presents the performance comparison under the point-wise missing pattern with a 10\% missing rate, evaluated across MAE, CRPS, and MAPE. 
Each subfigure compares the two configurations and highlights the performance gap between them. 
Specifically, we compute $\Delta = \text{Score}_{T=288,D=9} - \text{Score}_{T=24,D=207}$ as a diagnostic measure of performance change across the two configurations. 
Although the two settings differ in both horizon length and dimensionality, this comparison provides a diagnostic view of model stability under a substantial change in temporal context.

Across all three metrics, FADTI shows a smaller performance change across the two configurations than competing methods, suggesting better stability under this diagnostic change in horizon length and dimensionality.

\subsection{Sensitivity to Noise Distributions}
\label{sec:noise_sensitivity}

\begin{figure}[!t]
    \centering
    \includegraphics[width=\columnwidth]{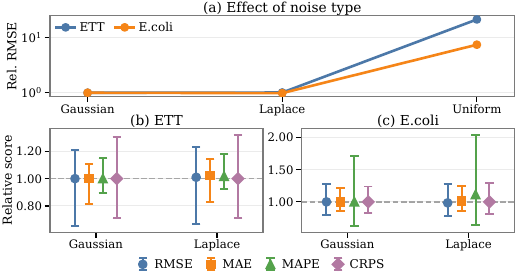}
    \caption{
    Noise sensitivity analysis. Errors are normalized to the Gaussian-noise setting; lower is better.
    }
    \label{fig:noise_effects}
\end{figure}

We replace Gaussian noise with unit-variance Laplace and Uniform noise to compare heavy-tailed and bounded alternatives under matched variance. 
Figure~\ref{fig:noise_effects} summarizes the relative error across metrics. 
Gaussian noise consistently yields the best performance on both datasets, while Laplace produces only minor degradation. 
In contrast, Uniform noise leads to substantially larger errors.

These results indicate that FADTI is sensitive to the choice of forward corruption distribution.
Gaussian perturbations match the DDPM training assumption and yield more stable score estimates, whereas distributional mismatch from Laplace or Uniform noise can degrade the learned reverse process.

\begin{table}[t]
\centering
\caption{Zero-shot imputation-to-forecasting results on E. coli. }
\label{tab:ecoli_impforecast}
\begin{tabular}{lcccc}
\toprule
Method & MAE & RMSE & MAPE & CRPS \\
\midrule
BRITS & 0.215$_{\scriptscriptstyle \pm 0.015}$ & 0.598$_{\scriptscriptstyle \pm 0.051}$ & 0.466$_{\scriptscriptstyle \pm 0.113}$ & 0.215$_{\scriptscriptstyle \pm 0.015}$ \\
SAITS & \underline{0.191$_{\scriptscriptstyle \pm 0.116}$} & \underline{0.317$_{\scriptscriptstyle \pm 0.206}$} & 0.682$_{\scriptscriptstyle \pm 0.540}$ & \underline{0.191$_{\scriptscriptstyle \pm 0.116}$} \\
CSDI & \textbf{\underline{0.098$_{\scriptscriptstyle \pm 0.007}$}} & \textbf{\underline{0.304$_{\scriptscriptstyle \pm 0.034}$}} & \textbf{\underline{0.256$_{\scriptscriptstyle \pm 0.036}$}} & \textbf{\underline{0.072$_{\scriptscriptstyle \pm 0.005}$}} \\
MTSCI & 0.303$_{\scriptscriptstyle \pm 0.009}$ & 0.901$_{\scriptscriptstyle \pm 0.030}$ & \underline{0.430$_{\scriptscriptstyle \pm 0.029}$} & 0.232$_{\scriptscriptstyle \pm 0.009}$ \\
\midrule 
FADTI & \underline{0.124$_{\scriptscriptstyle \pm 0.018}$} & \underline{0.375$_{\scriptscriptstyle \pm 0.058}$} & \underline{0.392$_{\scriptscriptstyle \pm 0.111}$} & \underline{0.090$_{\scriptscriptstyle \pm 0.013}$} \\
\bottomrule
\end{tabular}
\end{table}

\subsection{Zero-shot Imputation-to-Forecasting}
\label{sec:zero_shot_forecasting}

To further examine whether imputation models can generalize beyond missing-value reconstruction, we conduct a zero-shot imputation-to-forecasting experiment on E. coli. 
Specifically, each model is first trained under the standard point-wise imputation setting and is then directly evaluated on a forecasting-style mask without task-specific fine-tuning. 
During evaluation, the historical context contains point-wise missing values, while the future horizon is fully masked. 
Metrics are computed only over the masked forecast window.

Table~\ref{tab:ecoli_impforecast} reports the results.
FADTI achieves the second-best MAE, RMSE, and CRPS, remains competitive on MAPE, and consistently outperforms deterministic imputers such as BRITS and SAITS.
Although CSDI performs best in this zero-shot setting, FADTI is the strongest non-CSDI baseline and outperforms MTSCI across all metrics.
These results indicate that FADTI retains useful extrapolation ability for future block recovery despite being trained only for imputation.

\section{Conclusion}
We proposed FADTI, a conditional diffusion framework for multivariate time series imputation that combines frequency-informed modulation with feature-wise temporal modeling.
To address uncertainty handling and lack of frequency-domain inductive biases in existing methods, we introduced the FBP module, which flexibly incorporates Fourier-based transforms rooted in classical signal processing, including DFT, STFT, and FSST.
We instantiate FBP with all three transforms and, to the best of our knowledge, apply STFT and FSST to multivariate time series imputation for the first time.

Extensive experiments show that FADTI achieves state-of-the-art performance across diverse datasets and missing patterns, with particularly strong gains at high missing rates.
Ablation studies further confirm the importance of frequency-domain priors, implemented via FBP, for improving imputation accuracy.
Future work includes extensions to irregular time grids, adaptive spectral transforms, and downstream tasks such as forecasting and anomaly detection.

\bibliographystyle{IEEEtran}
\bibliography{bibfile}

@IEEEtranBSTCTL{FADTI:BSTcontrol,
  CTLuse_forced_etal       = {yes},
  CTLmax_names_forced_etal = {6},
  CTLnames_show_etal       = {1}
}

@article{cit:subgraphsurvey2026runze,
  author  = {Runze Li and Hanchen Wang and Ying Zhang and Wenjie Zhang},
  title   = {Algorithmic and Learning-Based Approaches to Subgraph Matching and Counting: {A} Survey},
  journal = {TechRxiv},
  year    = {2026},
}

@inproceedings{cit:knnjoin2023nimish,
  author    = {Nimish Ukey and Zhengyi Yang and Wenke Yang and Binghao Li and Runze Li},
  title     = {{kNN} Join for Dynamic High-Dimensional Data: {A} Parallel Approach},
  booktitle = {Proc. Australas. Database Conf. (ADC)},
  pages     = {3--16},
  year      = {2023},
}

@inproceedings{cit:ddpm2020ho,
  author       = {Jonathan Ho and
                  Ajay Jain and
                  Pieter Abbeel},
  title        = {Denoising Diffusion Probabilistic Models},
  booktitle    = {Adv. Neural Inf. Process. Syst. (NeurIPS)},
  year         = {2020},
}

@inproceedings{cit:fbp2024runze,
  author       = {Runze Yang and
                  Longbing Cao and
                  Jie Yang and
                  Jianxun Li},
  title        = {Rethinking Fourier Transform from {A} Basis Functions Perspective
                  for Long-term Time Series Forecasting},
  booktitle    = {Adv. Neural Inf. Process. Syst. (NeurIPS)},
  year         = {2024},
}

@article{cit:chen2024laplacian,
  title={Laplacian convolutional representation for traffic time series imputation},
  author={Chen, Xinyu and Cheng, Zhanhong and Cai, HanQin and Saunier, Nicolas and Sun, Lijun},
  journal={IEEE Trans. Knowl. Data Eng.},
  volume={36},
  number={11},
  pages={6490--6502},
  year={2024},
}

@inproceedings{cit:timesnet2023haixu,
  author       = {Haixu Wu and
                  Tengge Hu and
                  Yong Liu and
                  Hang Zhou and
                  Jianmin Wang and
                  Mingsheng Long},
  title        = {TimesNet: Temporal 2D-Variation Modeling for General Time Series Analysis},
  booktitle    = {Proc. Int. Conf. Learn. Represent. (ICLR)},
  year         = {2023},
}

@article{cit:mice2011stef,
  title={MICE: Multivariate Imputation by Chained Equations in R},
  author={Stef van Buuren and Karin G. M. Groothuis-Oudshoorn},
  journal={J. Stat. Softw.},
  year={2011},
  volume={45},
  pages={1--67},
}

@article{cit:knn2019ching,
  author       = {Ching{-}Hsue Cheng and
                  Chia{-}Pang Chan and
                  Yu{-}Jheng Sheu},
  title        = {A novel purity-based k nearest neighbors imputation method and its
                  application in financial distress prediction},
  journal      = {Eng. Appl. Artif. Intell.},
  volume       = {81},
  pages        = {283--299},
  year         = {2019},
}

@article{cit:cudre2020mind,
  title={Mind the gap},
  author={Cudr{\'e}-Mauroux, Philippe and Tymchenko, Zakhar and Lerner, Alberto},
  journal={Proc. VLDB Endow.},
  volume={13},
  number={5},
  pages={768--782},
  year={2020}
}

@inproceedings{cit:brits2018wei,
  author       = {Wei Cao and
                  Dong Wang and
                  Jian Li and
                  Hao Zhou and
                  Lei Li and
                  Yitan Li},
  title        = {{BRITS:} Bidirectional Recurrent Imputation for Time Series},
  booktitle    = {Adv. Neural Inf. Process. Syst. (NeurIPS)},
  pages        = {6776--6786},
  year         = {2018},
}

@inproceedings{cit:gan2018yonghong,
  author       = {Yonghong Luo and
                  Xiangrui Cai and
                  Ying Zhang and
                  Jun Xu and
                  Xiaojie Yuan},
  title        = {Multivariate Time Series Imputation with Generative Adversarial Networks},
  booktitle    = {Adv. Neural Inf. Process. Syst. (NeurIPS)},
  pages        = {1603--1614},
  year         = {2018},
}

@inproceedings{cit:naomi2019yukai,
  author       = {Yukai Liu and
                  Rose Yu and
                  Stephan Zheng and
                  Eric Zhan and
                  Yisong Yue},
  title        = {{NAOMI:} Non-Autoregressive Multiresolution Sequence Imputation},
  booktitle    = {Adv. Neural Inf. Process. Syst. (NeurIPS)},
  pages        = {11236--11246},
  year         = {2019},
}

@inproceedings{cit:csdi2021yusuke,
  author       = {Yusuke Tashiro and
                  Jiaming Song and
                  Yang Song and
                  Stefano Ermon},
  title        = {{CSDI:} Conditional Score-based Diffusion Models for Probabilistic
                  Time Series Imputation},
  booktitle    = {Adv. Neural Inf. Process. Syst. (NeurIPS)},
  pages        = {24804--24816},
  year         = {2021},
}

@inproceedings{cit:gnn2022andrea,
  author       = {Andrea Cini and
                  Ivan Marisca and
                  Cesare Alippi},
  title        = {Filling the G{\_}ap{\_}s: Multivariate Time Series Imputation by Graph
                  Neural Networks},
  booktitle    = {Proc. Int. Conf. Learn. Represent. (ICLR)},
  year         = {2022},
}

@article{cit:saits2023wenjie,
  author       = {Wenjie Du and
                  David C{\^{o}}t{\'{e}} and
                  Yan Liu},
  title        = {{SAITS:} Self-attention-based imputation for time series},
  journal      = {Expert Syst. Appl.},
  volume       = {219},
  pages        = {119619},
  year         = {2023},
}

@inproceedings{cit:imputeformer2024tong,
  author       = {Tong Nie and
                  Guoyang Qin and
                  Wei Ma and
                  Yuewen Mei and
                  Jian Sun},
  title        = {ImputeFormer: Low Rankness-Induced Transformers for Generalizable
                  Spatiotemporal Imputation},
  booktitle    = {Proc. ACM SIGKDD Int. Conf. Knowl. Discov. Data Min.},
  pages        = {2260--2271},
  year         = {2024},
}

@inproceedings{cit:pristi2023mingzhe,
  author       = {Mingzhe Liu and
                  Han Huang and
                  Hao Feng and
                  Leilei Sun and
                  Bowen Du and
                  Yanjie Fu},
  title        = {PriSTI: {A} Conditional Diffusion Framework for Spatiotemporal Imputation},
  booktitle    = {Proc. IEEE Int. Conf. Data Eng. (ICDE)},
  pages        = {1927--1939},
  year         = {2023},
}

@inproceedings{cit:gpvae2020vincent,
  author       = {Vincent Fortuin and
                  Dmitry Baranchuk and
                  Gunnar R{\"{a}}tsch and
                  Stephan Mandt},
  title        = {{GP-VAE:} Deep Probabilistic Time Series Imputation},
  booktitle    = {Proc. Int. Conf. Artif. Intell. Stat. (AISTATS)},
  volume       = {108},
  pages        = {1651--1661},
  year         = {2020},
}

@article{cit:vae2022ahmad,
  author       = {Ahmad Wisnu Mulyadi and
                  Eunji Jun and
                  Heung{-}Il Suk},
  title        = {Uncertainty-Aware Variational-Recurrent Imputation Network for Clinical
                  Time Series},
  journal      = {IEEE Trans. Cybern.},
  volume       = {52},
  number       = {9},
  pages        = {9684--9694},
  year         = {2022},
}

@article{cit:survey2024yiyuan,
  author       = {Yiyuan Yang and
                  Ming Jin and
                  Haomin Wen and
                  Chaoli Zhang and
                  Yuxuan Liang and
                  Lintao Ma and
                  Yi Wang and
                  Chenghao Liu and
                  Bin Yang and
                  Zenglin Xu and
                  Jiang Bian and
                  Shirui Pan and
                  Qingsong Wen},
  title        = {A Survey on Diffusion Models for Time Series and Spatio-Temporal Data},
  journal      = {CoRR},
  volume       = {abs/2404.18886},
  year         = {2024},
}

@article{cit:diffusion2023juan,
  author       = {Juan Miguel Lopez Alcaraz and
                  Nils Strodthoff},
  title        = {Diffusion-based Time Series Imputation and Forecasting with Structured
                  State Space Models},
  journal      = {Trans. Mach. Learn. Res.},
  volume       = {2023},
  year         = {2023},
}

@inproceedings{cit:csbi2023yu,
  author       = {Yu Chen and
                  Wei Deng and
                  Shikai Fang and
                  Fengpei Li and
                  Nicole Tianjiao Yang and
                  Yikai Zhang and
                  Kashif Rasul and
                  Shandian Zhe and
                  Anderson Schneider and
                  Yuriy Nevmyvaka},
  title        = {Provably Convergent Schr{\"{o}}dinger Bridge with Applications
                  to Probabilistic Time Series Imputation},
  booktitle    = {Proc. Int. Conf. Mach. Learn. (ICML)},
  volume       = {202},
  pages        = {4485--4513},
  year         = {2023},
}

@inproceedings{cit:wang2023observed,
  title={An observed value consistent diffusion model for imputing missing values in multivariate time series},
  author={Wang, Xu and Zhang, Hongbo and Wang, Pengkun and Zhang, Yudong and Wang, Binwu and Zhou, Zhengyang and Wang, Yang},
  booktitle={Proc. ACM SIGKDD Int. Conf. Knowl. Discov. Data Min.},
  pages={2409--2418},
  year={2023}
}

@inproceedings{cit:sadi2024zongyu,
  author       = {Zongyu Dai and
                  Emily J. Getzen and
                  Qi Long},
  title        = {{SADI:} Similarity-Aware Diffusion Model-Based Imputation for Incomplete
                  Temporal {EHR} Data},
  booktitle    = {Proc. Int. Conf. Artif. Intell. Stat. (AISTATS)},
  volume       = {238},
  pages        = {4195--4203},
  year         = {2024},
}

@inproceedings{cit:mtsci2024jianping,
  author       = {Jianping Zhou and
                  Junhao Li and
                  Guanjie Zheng and
                  Xinbing Wang and
                  Chenghu Zhou},
  title        = {{MTSCI:} {A} Conditional Diffusion Model for Multivariate Time Series
                  Consistent Imputation},
  booktitle    = {Proc. ACM Int. Conf. Inf. Knowl. Manag. (CIKM)},
  pages        = {3474--3483},
  year         = {2024},
}

@inproceedings{cit:fgti2024xin,
  author       = {Xinyu Yang and
                  Yu Sun and
                  Xiaojie Yuan and
                  Xinyang Chen},
  title        = {Frequency-aware Generative Models for Multivariate Time Series Imputation},
  booktitle    = {Adv. Neural Inf. Process. Syst. (NeurIPS)},
  year         = {2024},
}

@inproceedings{cit:iot2023xiao,
  author       = {Xiao Li and
                  Huan Li and
                  Harry Kai{-}Ho Chan and
                  Hua Lu and
                  Christian S. Jensen},
  title        = {Data Imputation for Sparse Radio Maps in Indoor Positioning},
  booktitle    = {Proc. IEEE Int. Conf. Data Eng. (ICDE)},
  pages        = {2235--2248},
  year         = {2023},
}

@article{cit:medical2012ibrahim,
  title={Missing data in clinical studies: issues and methods},
  author={Ibrahim, Joseph G and Chu, Haitao and Chen, Ming-Hui},
  journal={J. Clin. Oncol.},
  volume={30},
  number={26},
  pages={3297--3303},
  year={2012},
}

@article{cit:medical2025linglong,
  author       = {Linglong Qian and
                  Tao Wang and
                  Jun Wang and
                  Hugh Logan Ellis and
                  Robin Mitra and
                  Richard J. B. Dobson and
                  Zina M. Ibrahim},
  title        = {How Deep is your Guess? {A} Fresh Perspective on Deep Learning for
                  Medical Time-Series Imputation},
  journal={IEEE J. Biomed. Health Inform.},
  year={2025},
}

@article{cit:economic2008jushan,
title = {Forecasting economic time series using targeted predictors},
journal = {J. Econometrics},
volume = {146},
number = {2},
pages = {304--317},
year = {2008},
author = {Jushan Bai and Serena Ng},
}

@article{cit:transportation2023yongshun,
  author       = {Yongshun Gong and
                  Zhibin Li and
                  Jian Zhang and
                  Wei Liu and
                  Yilong Yin and
                  Yu Zheng},
  title        = {Missing Value Imputation for Multi-View Urban Statistical Data via
                  Spatial Correlation Learning},
  journal      = {IEEE Trans. Knowl. Data Eng.},
  volume       = {35},
  number       = {1},
  pages        = {686--698},
  year         = {2023},
}

@article{cit:stft2022sudhakar,
  author       = {Sudhakar Kumawat and
                  Manisha Verma and
                  Yuta Nakashima and
                  Shanmuganathan Raman},
  title        = {Depthwise Spatio-Temporal {STFT} Convolutional Neural Networks for
                  Human Action Recognition},
  journal      = {IEEE Trans. Pattern Anal. Mach. Intell.},
  volume       = {44},
  number       = {9},
  pages        = {4839--4851},
  year         = {2022},
}

@article{cit:fourier1977allen,
  title={A unified approach to short-time Fourier analysis and synthesis},
  author={Allen, Jont B and Rabiner, Lawrence R},
  journal={Proc. IEEE},
  volume={65},
  number={11},
  pages={1558--1564},
  year={1977},
}

@article{cit:frft2020jun,
  author       = {Jun Shi and
                  Jiabin Zheng and
                  Xiaoping Liu and
                  Wei Xiang and
                  Qinyu Zhang},
  title        = {Novel Short-Time Fractional Fourier Transform: Theory, Implementation,
                  and Applications},
  journal      = {IEEE Trans. Signal Process.},
  volume       = {68},
  pages        = {3280--3295},
  year         = {2020},
}

@article{cit:frft1980victor,
  title={The fractional order Fourier transform and its application to quantum mechanics},
  author={Victor Namias},
  journal={IMA J. Appl. Math.},
  volume={25},
  number={3},
  pages={241--265},
  year={1980},
}

@article{cit:fsst2013Gaurav,
  author       = {Gaurav Thakur and
                  Eugene Brevdo and
                  Neven S. Fuckar and
                  Hau{-}Tieng Wu},
  title        = {The Synchrosqueezing algorithm for time-varying spectral analysis:
                  Robustness properties and new paleoclimate applications},
  journal      = {Signal Process.},
  volume       = {93},
  number       = {5},
  pages        = {1079--1094},
  year         = {2013},
}

@article{cit:sst2011ingrid,
  title={Synchrosqueezed wavelet transforms: An empirical mode decomposition-like tool},
  author={Ingrid Daubechies and Jianfeng Lu and Hau‐Tieng Wu},
  journal={Appl. Comput. Harmon. Anal.},
  year={2011},
  volume={30},
  number={2},
  pages={243--261},
}

@inproceedings{cit:timemixer++2025shiyu,
  author       = {Shiyu Wang and
                  Jiawei Li and
                  Xiaoming Shi and
                  Zhou Ye and
                  Baichuan Mo and
                  Wenze Lin and
                  Shengtong Ju and
                  Zhixuan Chu and
                  Ming Jin},
  title        = {TimeMixer++: {A} General Time Series Pattern Machine for Universal
                  Predictive Analysis},
  booktitle    = {Proc. Int. Conf. Learn. Represent. (ICLR)},
  year         = {2025},
}

@inproceedings{cit:timemixer2024shiyu,
  author       = {Shiyu Wang and
                  Haixu Wu and
                  Xiaoming Shi and
                  Tengge Hu and
                  Huakun Luo and
                  Lintao Ma and
                  James Y. Zhang and
                  Jun Zhou},
  title        = {TimeMixer: Decomposable Multiscale Mixing for Time Series Forecasting},
  booktitle    = {Proc. Int. Conf. Learn. Represent. (ICLR)},
  year         = {2024},
}

@inproceedings{cit:informer2021haoyi,
  author       = {Haoyi Zhou and
                  Shanghang Zhang and
                  Jieqi Peng and
                  Shuai Zhang and
                  Jianxin Li and
                  Hui Xiong and
                  Wancai Zhang},
  title        = {Informer: Beyond Efficient Transformer for Long Sequence Time-Series
                  Forecasting},
  booktitle    = {Proc. AAAI Conf. Artif. Intell.},
  pages        = {11106--11115},
  year         = {2021},
}

@article{cit:ecoli2024jean,
  title={Deep model predictive control of gene expression in thousands of single cells},
  author={Lugagne, Jean-Baptiste and Blassick, Caroline M and Dunlop, Mary J},
  journal={Nat. Commun.},
  volume={15},
  number={1},
  pages={2148},
  year={2024},
}

@inproceedings{cit:metrla2018yaguang,
  author       = {Yaguang Li and
                  Rose Yu and
                  Cyrus Shahabi and
                  Yan Liu},
  title        = {Diffusion Convolutional Recurrent Neural Network: Data-Driven Traffic
                  Forecasting},
  booktitle    = {Proc. Int. Conf. Learn. Represent. (ICLR)},
  year         = {2018},
}

@inproceedings{cit:tslanet2024emadeldeen,
  author       = {Emadeldeen Eldele and
                  Mohamed Ragab and
                  Zhenghua Chen and
                  Min Wu and
                  Xiaoli Li},
  title        = {TSLANet: Rethinking Transformers for Time Series Representation Learning},
  booktitle    = {Proc. Int. Conf. Mach. Learn. (ICML)},
  year         = {2024},
}

@article{cit:attention2025cizheng,
  author       = {Caizheng Liu and
                  Zhengyu Zhu and
                  Wanming Hao and
                  Gangcan Sun},
  title        = {Heterogeneous multivariate time series imputation by transformer model
                  with missing position encoding},
  journal      = {Expert Syst. Appl.},
  volume       = {271},
  pages        = {126435},
  year         = {2025},
}

@article{cit:dsttn2024xusheng,
  title={Cross-modal missing time-series imputation using dense spatio-temporal transformer nets},
  author={Xusheng Qian and Teng Zhang and Meng Miao and Gaojun Xu and Xuancheng Zhang and Wenwu Yu and Duxin Chen},
  journal={Math. Biosci. Eng.},
  volume={21},
  number={4},
  pages={4989--5006},
  year={2024}
}

@inproceedings{cit:lai2024rectsi,
  author       = {Zhichen Lai and
                  Dalin Zhang and
                  Huan Li and
                  Dongxiang Zhang and
                  Hua Lu and
                  Christian S. Jensen},
  title        = {ReCTSi: Resource-efficient Correlated Time Series Imputation via Decoupled
                  Pattern Learning and Completeness-aware Attentions},
  booktitle    = {Proc. ACM SIGKDD Int. Conf. Knowl. Discov. Data Min.},
  pages        = {1474--1483},
  year         = {2024},
}

@article{cit:dft1965james,
  title={An algorithm for the machine calculation of complex Fourier series},
  author={James W. Cooley and John W. Tukey},
  journal={Math. Comput.},
  year={1965},
  volume={19},
  pages={297--301},
}

@inproceedings{wang2025optimal,
  title={Optimal transport for time series imputation},
  author={Wang, Hao and Li, Haoxuan and Chen, Xu and Gong, Mingming and Chen, Zhichao and others},
  booktitle={Proc. Int. Conf. Learn. Represent. (ICLR)},
  year={2025}
}

@inproceedings{cit:ssdts2025Hongfan,
  author       = {Hongfan Gao and
                  Wangmeng Shen and
                  Xiangfei Qiu and
                  Ronghui Xu and
                  Bin Yang and
                  Jilin Hu},
  title        = {{SSD-TS:} Exploring the Potential of Linear State Space Models for
                  Diffusion Models in Time Series Imputation},
  booktitle    = {Proc. ACM SIGKDD Int. Conf. Knowl. Discov. Data Min.},
  pages        = {649--660},
  year         = {2025},
}

\end{document}